\documentclass[lettersize,journal]{IEEEtran}
\usepackage{amsmath}
\usepackage{amsfonts}
\usepackage{algorithm}
\usepackage{algpseudocode}
\usepackage{array}
\usepackage[caption=false,font=normalsize,labelfont=sf,textfont=sf]{subfig}
\usepackage{textcomp}
\usepackage{stfloats}
\usepackage{url}
\usepackage{verbatim}
\usepackage{graphicx}
\usepackage{cite}
\usepackage{xcolor}
\usepackage{bm}
\usepackage{tabularx}
\usepackage{booktabs} 
\usepackage{multirow} 
\usepackage{soul}
\usepackage{amssymb}
\usepackage{threeparttable}

\newtheorem{proposition}{Proposition}

\begin{document}

\title{NiSNN-A: Non-iterative Spiking Neural Networks with Attention with Application to Motor Imagery EEG Classification}

\author{Chuhan Zhang, Wei Pan, Cosimo Della Santina
\thanks{This work is supported by the TU Delft AI Labs programme.}
\thanks{Chuhan Zhang, Cosimo Della Santina are with the Department of Cognitive Robotics, Faculty of Mechanical Engineering, Delft University of Technology, Delft, Netherlands (e-mail: C.Zhang-8@tudelft.nl; C.DellaSantina@tudelft.nl).}
\thanks{Wei Pan is with the Department of Computer Science, The University of Manchester, Manchester, United Kingdom (e-mail: wei.pan@manchester.ac.uk).}
}

\maketitle

\begin{abstract}
Motor imagery, an important category in electroencephalogram (EEG) research, often intersects with scenarios demanding low energy consumption, such as portable medical devices and isolated environment operations. Traditional deep learning algorithms, despite their effectiveness, are characterized by significant computational demands accompanied by high energy usage. As an alternative, spiking neural networks (SNNs), inspired by the biological functions of the brain, emerge as a promising energy-efficient solution. However, SNNs typically exhibit lower accuracy than their counterpart convolutional neural networks (CNNs). Although attention mechanisms successfully increase network accuracy by focusing on relevant features, their integration in the SNN framework remains an open question. In this work, we combine the SNN and the attention mechanisms for the EEG classification, aiming to improve precision and reduce energy consumption. To this end, we first propose a Non-iterative Leaky Integrate-and-Fire (NiLIF) neuron model, overcoming the gradient issues in traditional Spiking Neural Networks (SNNs) that use Iterative LIF neurons for long time steps. Then, we introduce the sequence-based attention mechanisms to refine the feature map. We evaluated the proposed Non-iterative SNN with Attention (NiSNN-A) model on two motor imagery EEG datasets, OpenBMI and BCIC IV 2a. Experiment results demonstrate that 1) our model outperforms other SNN models by achieving higher accuracy, 2) our model increases energy efficiency compared to the counterpart CNN models (i.e., by 2.13 times) while maintaining comparable accuracy.
\end{abstract}

\begin{IEEEkeywords}
Spiking neural networks, Attention mechanism, Motor imagery, EEG classification.
\end{IEEEkeywords}

\section{Introduction}
\label{Introduction}
\IEEEPARstart{E}{lectroencephalograms} (EEGs), whether captured through non-invasive electrodes on the scalp or directly via invasive devices, are the cornerstone of the rapidly evolving domain of Brain-Computer Interfaces (BCI). Therefore, accurate classification of EEG signals has attracted substantial attention over the years - with applications ranging from advanced neurorehabilitation techniques \cite{tsiamalou2022eeg} to diagnostics and real-time health monitoring \cite{del2023single}. Within the realm of BCI, motor imagery (MI) holds a distinctive place \cite{singh2021comprehensive}. MI refers to the mental imagination of specific movements by subjects, leading to the generation of distinct EEG patterns. Accurately classifying these EEG signals opens up application possibilities in advanced fields like robot control and assistive technologies \cite{zhang2021survey}.

Recently, deep learning (DL) methods such as Convolutional Neural Networks (CNNs) \cite{rajwal2023convolutional}, Recurrent Neural Networks (RNNs) \cite{alhagry2017emotion}, and Transformers \cite{song2022eeg} have received increasing interest as a mean for classifying EEG signals \cite{kwon2020subjectindependent, bang2022spatiospectral, zhang2022explainable}.
At present, in addition to conventional CNNs or RNNs, DL of EEG incorporates many other technologies \cite{ju2023tensorcspnet, liu2020parallel, eldele2021attention, liao2023convolutional, zhang2022improving}. Attention mechanisms \cite{vaswani2017attention}, inspired by human cognitive processes, enhance the model's focus on relevant features, facilitating more efficient and accurate feature extraction \cite{li2020efficient}. Attention mechanisms have been successfully applied to EEG classification. More details are presented in Section~\ref{related_work.attention}. However, all of these methods suffer from the current limitation of high energy consumption, which poses a significant barrier to deployment in low energy scenarios such as edge devices for healthcare \cite{abirami2020energy} or robot control \cite{huang2021review}.

In this work, we investigate the use of Spiking Neural Networks (SNNs) for EEG classification. This technique mimics how biological neurons operate, allowing it to be used to interpret natural neuronal signals \cite{wozniak2020deep}. SNNs also offer a promising avenue for reducing energy consumption due to their event-based nature \cite{schuman2022opportunities,liu2023spiking,siddiqi2023lightweight}, which is attractive in view of edge applications. More details on the state-of-the-art of SNN in EEG classification are provided in Section~\ref{related_work.snn}. 
We propose a novel integration of SNNs and the attention mechanism, especially for EEG classification. 
At the core of our approach sits a newly proposed Non-iterative Leaky Integrate-and-Fire (NiLIF) neuron, which approximates the neural dynamics of the biological LIF and mitigates the gradient problem by avoiding long-term dependencies. The second methodological contribution is the sequence-based attention model for EEG data, which can simultaneously obtain the attention scores of feature maps. The Non-iterative SNN with attention (NiSNN-A) models boosts both the efficiency and accuracy of the execution process.
It is worth noting that some work has investigated attention SNNs, which, however, specifically targeted computer vision \cite{yao2023attention}. As we show in our experimental comparison, this method is not suitable for classifying long-term data such as EEG. 

The contributions of this paper can be summarized as follows:
\begin{enumerate}
    \item We propose a novel non-iterative LIF neuron model for SNNs, which eliminates the gradient problem caused by long-term dependencies while retaining the biology-inspired temporal properties of the LIF model.
    \item We introduce a sequence-based attention mechanism for SNNs, improving the classification accuracy.
    \item We show the combination of NiSNNs with attention mechanisms for motor imagery EEG classification, simultaneously achieving high accuracy and reducing energy consumption. 
\end{enumerate}

The rest of the paper is organized as follows. Section~\ref{Methods} introduces our proposed NiSNN-A models. Section~\ref{Experiments} gives the experiment details. The results and discussion are conducted in Section~\ref{ResultAndDiscussion}. Section~\ref{Conclusion} concludes the work.

\section{Related Work}
\label{related_work}

In this section, the related works are introduced. Section~\ref{related_work.deep_learning} illustrates the application of deep learning techniques specific to EEG signal processing. Subsequently, Section~\ref{related_work.attention} illustrates the works with attention mechanisms. Finally, the applications of SNN in EEG classification are described in Section~\ref{related_work.snn}.

\subsection{Deep learning methods for EEG}
\label{related_work.deep_learning}
In recent years, integrating deep learning techniques into EEG classification has gained significant traction \cite{al2021deep}. CNN stands out among these techniques, particularly due to its ability to identify spatial-temporal patterns within the complex, multi-dimensional EEG data \cite{cui2023eegbased,dong2017mixed,wang2023ssgcnet}. For instance, a temporal-spatial CNN was employed for EEG classification in \cite{dose2018end, li2021temporal}. The work in \cite{lawhern2018eegnet} introduced EEGNet, which integrates a separable convolutional layer following the temporal and spatial modules. Enhancing the CNN architecture, \cite{huang2021electroencephalogram, humayun2019end} incorporated residual blocks for classification. Another noteworthy approach is presented in \cite{autthasan2021min2net}, where an autoencoder is built upon the CNN framework. Furthermore, this work adopted a subject-independent training paradigm, emphasizing its scalability on varied EEG data from different subjects. Apart from CNNs, Long Short-Term Memory (LSTM) networks have also been recognized for EEG classification due to their inherent ability to process time sequences effectively \cite{ma2018improving, zhao2021plug,xia2023mulhita}. These methods are limited in high energy consumption, making them difficult to use on some edge devices with low-energy requirements.

\subsection{Attention mechanism}
\label{related_work.attention}
The Transformer architecture and attention mechanism, originally introduced in \cite{vaswani2017attention} for natural language processing tasks, have seen increasing adoption in many other machine learning tasks \cite{brown2020language, achiam2023gpt}. These methods are now being explored in the field of EEG classification, given their ability to handle time sequence data. The attention mechanism is particularly important for EEG data analysis. It holds the potential to enhance classification accuracy and emphasizes specific segments of the data, offering deeper insights into EEG signal characteristics.
Various attention models have emerged in the field of EEG classification. For instance, \cite{liu2020parallel} presents a spatial and temporal attention model integrated with CNN. This approach leverages two distinct CNN modules to derive spatial and temporal attention scores separately, subsequently using four convolutional layers for classification. Meanwhile, \cite{zhang2023multi} applies a multi-head attention module, as described in \cite{vaswani2017attention}, combined with five convolutional layers to classify EEG signals. Direct applications of the Transformer attention structures from \cite{vaswani2017attention} for EEG classification are evident in \cite{wang2022cnn, wen2022new}. Beyond solely leveraging CNN and attention mechanisms, some studies have integrated additional techniques. For instance, \cite{luo2023shallow} introduces a mirrored input approach combined with an attention model that operates across each data record. In a more intricate approach, \cite{ma2022novel} deploys spatial, spectral, and temporal Transformers, each catering to a different input data type. A popular EEG processing methodology, time-frequency Common Spatial Pattern (TFCSP), is highlighted in \cite{zhang2023self}. This method intertwines a two-layer CNN and an attention model, feeding their concatenated outputs into a classifier. A notable trend is the use of global attention in conjunction with three sequential models, as seen in \cite{wang2021residual, li2023parallel, fan2021bilinear}. These works employ three sequential attention models, each dedicated to a single dimension. By utilizing Global Average Pooling (GAP), they efficiently diminish unrelated dimensions and consolidate the attention scores across the three output dimensions. These models effectively achieved the goal of EEG signal classification. However, these methods have the limitation of high energy consumption due to the use of attention CNN, making them difficult to use on some edge devices with low-energy requirements. Also, they lack a special focus on data from different channels and different time areas, which is important in EEG data.
Attention mechanisms for SNNs specific to computer vision tasks \cite{yao2023attention} are discussed in Sec.~\ref{Methods.attention}. 
\cite{cai2023spatial} also proposed a vision-based SNN attention mechanism, in which Spatial-Channel-Temporal-Fused attention works at each time step and layer of the spike trains, providing strong robustness. However, \cite{cai2023spatial} requires attention calculation at each time step in an iterative manner, which takes a long time to execute due to a large number of loops.

\subsection{Spiking neural networks}
\label{related_work.snn}

In recent years, SNNs have garnered increasing attention within the neural computing community with broad applications such as computer vision and robot control \cite{barton2023proposal, liu2020unsupervised, liu2020deep, safa2023improving, liu2023spiking}. Their growing significance can be attributed to their closer resemblance to biological neural systems than CNNs. SNNs, by simulating the discrete, spike-based communication found in actual neurons, promise enhanced efficiency and energy savings. However, this bio-inspired approach comes with challenges, particularly during training. Gradient backpropagation, a staple in training CNNs, presents difficulties for SNNs due to their non-differentiable spiking nature. 
To address this, various training methods have been proposed \cite{dampfhoffer2023backpropagation, zhang2022tuning}: from leveraging evolutionary algorithms to adjust synaptic weights \cite{schliebs2013evolving}, to employing biologically-inspired synaptic update rules like Spiking-Time Dependent Plasticity (STDP) \cite{kheradpisheh2018stdp}. Some researchers have also explored converting a well-trained CNN into their SNN counterparts \cite{rueckauer2017conversion}, while others have advocated for using surrogate gradients for continuing backpropagation \cite{wu2018spatio}. Given the biology-inspired and efficient attributes of SNNs, several works have successfully applied SNNs in the domain of EEG signal classification, showcasing their potential in real-world applications. For example, \cite{antelis2020spiking} employed Particle Swarm Optimization as an evolutionary algorithm for weight updates, combined with an unsupervised classifier like K Nearest Neighbours and Multilayer Perceptron; however, their approach was not end-to-end and involved manual feature extraction. Studies that utilized STDP include \cite{luo2020eeg}, which integrated manual feature extraction methods like Fast Fourier Transform and Discrete Wavelet Transform with a 3D SNN reservoir and supervised classifiers. Similarly, \cite{kasabov2015spiking} adopted an unsupervised learning framework, implementing a 3D SNN reservoir model.  \cite{wu2023improving} combined the 3D reservoir with a Support Vector Machine as the classifier. Highlighting conversion techniques, \cite{yan2021energy} explored a tree structure, demonstrating the energy efficiency benefits of SNNs through a CNN-to-SNN conversion, while another work by \cite{yan2022eeg} utilized Power Spectral Density for feature extraction before such a conversion. 
Back propagation-like methodologies also found their adaptation in SNNs and EEG classification realm with work \cite{ghosh2007improved} using SpikeProp \cite{bohte2002error}. 
SpikeProp uses solutions of dynamics equations to represent the membrane potentials, and uses backpropagation to calculate the timing of output spikes. However, SpikeProp considers a single input spike and output spike while our NiLIF model allows arbitrary inputs and multiple output spikes. In particular, the inputs to the NiLIF neuron can even be continuous values.
\cite{liao2023convolutional} were notably the first to employ directly-trained SNNs for EEG signal classification. However, these methods have limitations when deepening the networks and seeking to learn complex representations. 
There are also some works using parallel techniques in SNNs \cite{fang2024parallel} considering the membrane potential without a reset mechanism, which is different from our method.

\begin{figure}
	\centering
    \subfloat[Iterative LIF neuron\label{fig_lif_neuron}]{
        \includegraphics[width=0.49\linewidth]{ 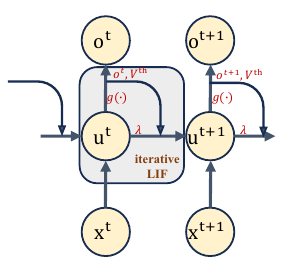}
    }
    \subfloat[Non-iterative LIF neuron\label{fig_non_lif_neuron}]{%
        \includegraphics[width=0.49\linewidth]{ 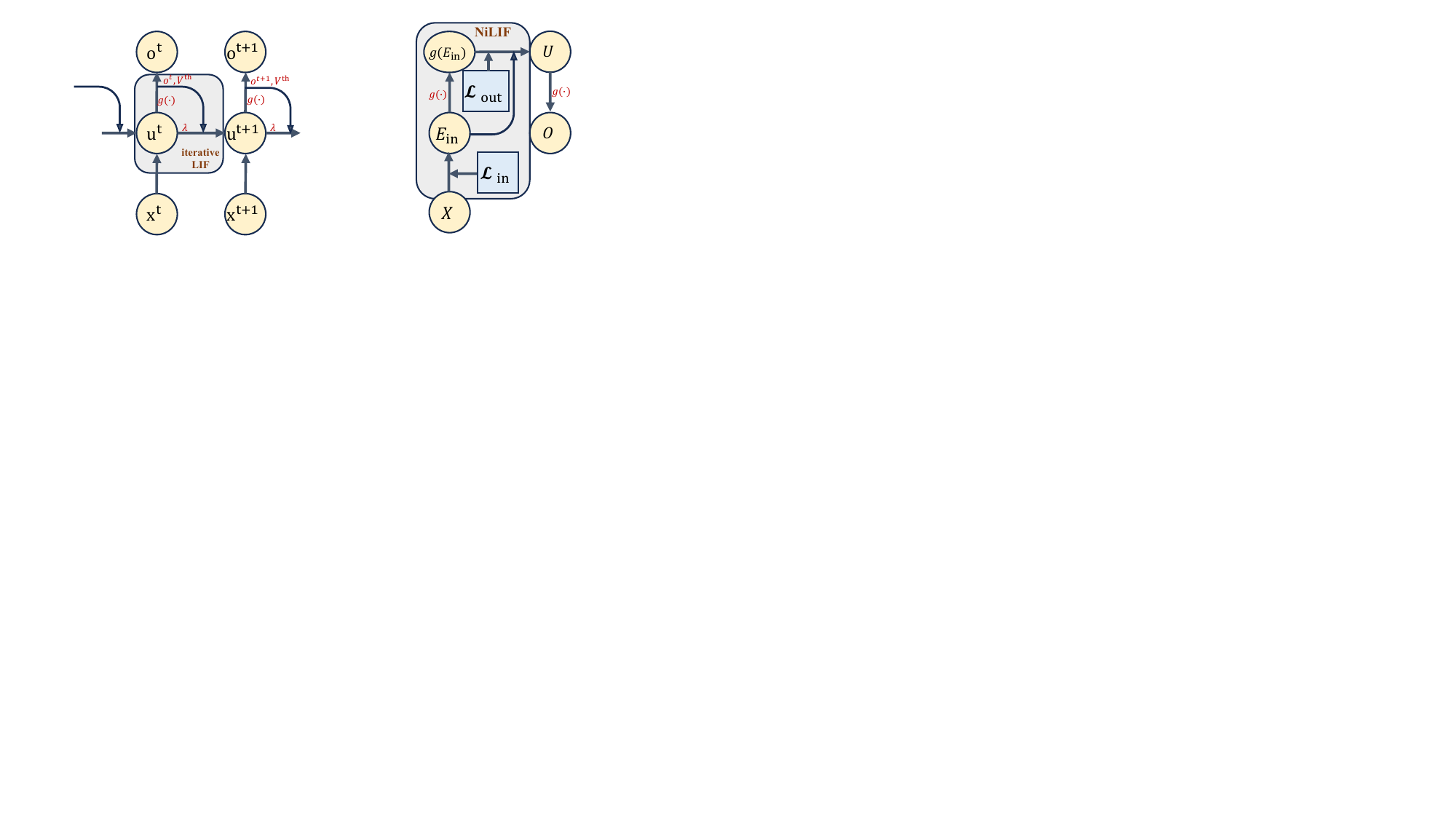}
    }
	\caption{The Iterative LIF neuron model and the NiLIF neuron model. (a) Iterative LIF neuron model. The membrane potential $u^{\text{t}+1}$ is computed recurrently, with each time step depends on its previous state $u^{\text{t}}$ and the output spikes $o^{\text{t}}$ from the preceding time step. (b) The NiLIF neuron model. The matrix $\bm{\mathcal{L}}_{\mathrm{in}}$ is used to calculate the input stimulus and accumulation processes, and the matrix $\bm{\mathcal{L}}_{\mathrm{out}}$ is used to approximate the output spikes. The final membrane potential $U$ and output spikes $O$ are obtained simultaneously rather than iteratively.}\label{fig_neuron_comparison}
\end{figure}

\section{Methods}
\label{Methods}

In this section, we introduce the novel NiLIF neuron in Section~\ref{Methods.lif}. Subsequently, the proposed attention models are delineated in Section~\ref{Methods.attention}. Finally, \ref{Methods.network} describes the network architecture of the proposed NiSNN-A.

\subsection{LIF neuron}
\label{Methods.lif}

Neurons serve as the fundamental components of neural networks. In this section, the Iterative LIF neuron model is presented in Section~\ref{Methods.lif.iterative}. Then, our proposed NiLIF neuron model is detailed in Section~\ref{Methods.lif.noniterative}. Finally, the comparisons are discussed in Section~\ref{Methods.lif.comparison}

\subsubsection{Background: Iterative LIF neuron model}
\label{Methods.lif.iterative}
In the SNN community, the Leaky Integrate-and-Fire (LIF) neuron model is widely used. It strikes a balance between simplicity and biologically inspired characteristics. The LIF model can be described using these equations:
\begin{equation}
\label{formula_lif}
\begin{aligned}
& \textbf{membrane potential:} \\
& \hspace{1.5em} \left\{ 
        \begin{array}{ll}
        \tau \frac{\mathrm{d} u(t_{\mathrm{c}})}{\mathrm{d} t_{\mathrm{c}}} = -u(t_{\mathrm{c}}) + wx(t_{\mathrm{c}}), \text{ if } u(t_{\mathrm{c}}) \leq V_{\mathrm{th}}, \\
        
        \lim\limits_{\Delta \to 0^{+}}u({t_{\mathrm{c}}}+\Delta) = u_{\mathrm{res}}, \text{ if } u(t_{\mathrm{c}}) > V_{\mathrm{th}},
        \end{array}\right. \\
& \hspace{1.5em} u_{\mathrm{res}} =
\left\{
    \begin{array}{ll}
        u({t_{\mathrm{c}}})-V_{\mathrm{th}}, & \text{ with soft reset mechanism,} \\
        u_{\mathrm{r}}, & \text{ with hard reset mechanism,}
    \end{array}
\right.\\
& \textbf{spike generation:} \\
& \hspace{1.5em} o(t_{\mathrm{c}}) = g(u_\mathrm{c}), \\
& \hspace{1.5em} g(a) = \left\{ 
      \begin{array}{ll}
        0, & \text{if } a \leq V_{\mathrm{th}}, \\
        1, & \text{if } a > V_{\mathrm{th}},
      \end{array} 
    \right. 
\end{aligned}
\end{equation}
where $\tau \in \mathbb{R}$ is the membrane time constant and $u(t_{\mathrm{c}}) \in \mathbb{R}$ represents the neuron's membrane potential at continuous time ${t_{\mathrm{c}}} \in \mathbb{R}$. $wx(t_{\mathrm{c}}) \in \mathbb{R}$ is the input stimulus at time ${t_{\mathrm{c}}} \in \mathbb{R}$, represented as the weighted input to the present layer in the neural network. $w$ is the trainable parameter. $V_{\mathrm{th}} \in \mathbb{R}$ is the membrane potential threshold. Specifically, when the membrane potential exceeds the threshold $V_{\mathrm{th}}$, a spike is produced. When the membrane potential remains below the threshold, no spike is generated. After generating a spike, the membrane potential decreases to a reset potential $u_{\mathrm{res}} \in \mathbb{R}$. Two types of reset mechanisms deal with the membrane potential after generating spikes: soft and hard reset. The soft reset mechanism resets the membrane potential by reducing the threshold potential $V_{\mathrm{th}}$; while the hard reset mechanism resets the membrane potential to a defined potential value $u_{\mathrm{r}} \in \mathbb{R}$.
$o(t_{\mathrm{c}}) \in \mathbb{R}$ represents the output spike at time $t_{\mathrm{c}}$. The function $g(\cdot)$ is the Heaviside step function, which describes the spike firing process.

To adapt to the requirements of backpropagation in neural networks, an Iterative LIF neuron model \cite{wu2018spatio} with the soft reset mechanism is introduced:
\begin{equation}
\begin{aligned}
    \label{formula_iterative_lif}
    u^{\mathrm{t+1}} &= \lambda ( u^{\mathrm{t}} - V_{\mathrm{th}} o^{\mathrm{t}}) + wx^{\mathrm{t}}, \\
    o^{\mathrm{t}} &= g(u^{\mathrm{t}}),
\end{aligned}
\end{equation}
where $\lambda \in \mathbb{R}$ denotes the decay rate of the membrane potential. We use $u^{\mathrm{t}}$ to represent $u(t_{\mathrm{c}}=t)$ where $t \in \mathbb{N}$ represents the discrete time step in the Iterative LIF neuron model. Similarly, $o^{\mathrm{t}}$ means $o(t_{\mathrm{c}}=t)$ and $x^{\mathrm{t}}$ means $x(t_{\mathrm{c}}=t)$. In this way, the membrane potential updates step by step, recurrently making the LIF neuron dynamics trainable by a network. However, the Heaviside step function $g(\cdot)$ in the firing process makes it non-differentiable. This characteristic brings challenges for gradient backpropagation. To address this limitation, surrogate functions have been proposed \cite{neftci2019surrogate}. Nowadays, the Sigmoid functions are commonly employed as surrogate functions due to their capability to emulate the spike firing process, especially when associated with a high value of $\alpha$: $\mathrm{Sigmoid}(x) = \frac{1}{1+e^{- \alpha x}}$. Therefore, the gradient backpropagation in the Iterative LIF with the Sigmoid functions can be described as:
\begin{equation}
    \frac{\partial o}{\partial u} = \frac{\partial \mathrm{Sigmoid}(u)}{\partial u} = \mathrm{Sigmoid}(u)(1-\mathrm{Sigmoid}(u))\alpha.
\end{equation}

\subsubsection{Non-iterative LIF neuron model} 
\label{Methods.lif.noniterative}

As described in \eqref{formula_iterative_lif}, the membrane potential $u^{\mathrm{t+1}}$ in Iterative LIF depends on the output $o^{\mathrm{t}}$ and membrane potential $u^{\mathrm{t}}$ of the preceding time steps, which introduces complex neuron dynamics and long-term dependencies in gradient propagation \cite{hochreiter2001gradient}. Therefore, for cases with long time steps, the gradient problem caused by long-term dependencies hinders the model's high performance. Thus, we propose a Non-iterative LIF (NiLIF) model designed to approximate the neuron dynamics and avoid long-term dependencies.

The LIF model has input stimulation, accumulation, and firing processes. To achieve these functions, the core idea behind the NiLIF model is first to consider the stimulation and accumulation effects over all time steps, then approximate the firing process with maximum sparsity, and finally obtain the approximate neuron dynamics without long-term dependencies.

The effects on time step $t$ of the input stimulation without firing at time step $i$ can be computed by solving the different equation \eqref{formula_lif}: $u^{\mathrm{t}} = wx^{\mathrm{i}} e^{-\frac{t-i}{\tau} {\delta t}} $, where $\delta t$ is the fixed time increment used to discretize the continuous time domain. We utilize the function $\mathcal{L}(\cdot)$ to represent the leaky component as $\mathcal{L}(t) = e^{-\frac{t}{\tau}{\delta t}}$. Therefore, the effects of all previous time steps on time step $t$ is expressed as:
\begin{equation}
\label{formula_lif_solution2}
     \epsilon_{\mathrm{in}}^{\mathrm{t}} = \sum^{t}_{i=0} wx^{\mathrm i} {\mathcal{L}}(t-i),
\end{equation}
where $\epsilon_{\mathrm{in}}^{\mathrm{t}}$ represents the input stimulus and accumulation effects without firing. Naturally, \eqref{formula_lif_solution2} can be represented using matrix format:
\begin{equation}
\label{formula.5}
    E_{\mathrm{in}} = X \bm{\mathcal{L}}_{\mathrm{in}},
\end{equation}
where $E_{\mathrm{in}} \in \mathbb{R}^{1 \times (t_{\mathrm{n}}+1)}$ is defined as the effect matrix, $X \in \mathbb{R}^{1 \times (t_{\mathrm{n}}+1)}$ is defined as the weighted input matrix, and $\bm{\mathcal{L}}_{\mathrm{in}} \in \mathbb{R}^{(t_{\text{n}}+1) \times (t_{\text{n}}+1)}$ is defined as the leaky matrix:
\begin{equation}
\begin{aligned}
\label{formula.6}
& E_{\mathrm{in}} =
\begin{bmatrix}
    \epsilon_{\mathrm{in}}^0 & \epsilon_{\mathrm{in}}^1 & \hdots & \epsilon_{\mathrm{in}}^{\mathrm{t_{\text{n}}}}
\end{bmatrix}, \\
& X = 
\begin{bmatrix}
 wx^0 & wx^1 & \hdots & wx^{\mathrm t_{\text{n}}}
\end{bmatrix}, \\
& \bm{\mathcal{L}}_{\mathrm{in}} =
\begin{bmatrix}
1 & \mathcal{L}(1) & \hdots & \mathcal{L}(t_{\text{n}}) \\
0 & 1 & \hdots & \mathcal{L}(t_{\text{n}}-1) \\
\vdots & \vdots & \ddots & \vdots \\
0 & 0 & \hdots & 1
\end{bmatrix}.
\end{aligned}
\end{equation}
Therefore, the stimulus and accumulation dynamics of the LIF neuron are represented.

Additionally, we approximate the firing dynamics of the LIF neuron. We propose to use the term $\epsilon_{\mathrm{out}}$ to represent the impact of output spikes from the previous time steps on the current time step t:
\begin{equation}
\label{formula_lif_solution3}
    \epsilon_{\mathrm{out}}^{\mathrm{t}}= \sum^{t-1}_{i=0} o^{\mathrm{i}} V_{\mathrm{th}} {\mathcal{L}}(t-i-1),
\end{equation}
where $o^{\mathrm{i}} V_{\mathrm{th}}$ represents the soft reset mechanism. When no spike is generated ($o^{\mathrm{i}}=0$), $o^{\mathrm{i}} V_{\mathrm{th}}$ is 0, which means no output spike effect at this time step $i$. When a spike is generated at time step $i$ ($o^{i}=1$), the membrane potential in the next step will decrease by the threshold $V_{\mathrm{th}}$, which will affect the current membrane potential with a leakage coefficient $\mathcal{L}(t-i-1)$. \eqref{formula_lif_solution3} can be represented using matrix format:
\begin{equation}
\label{formula.7}
    E_{\mathrm{out}} = O \bm{\mathcal{L}}_{\mathrm{out}},
\end{equation}
where  $E_{\mathrm{out}} \in \mathbb{R}^{1 \times (t_{\mathrm{n}}+1)}$ is defined as the output spikes effect matrix, $O \in \mathbb{R}^{1 \times (t_{\mathrm{n}}+1)}$ is defined as the output spikes matrix, and $\bm{\mathcal{L}}_{\mathrm{out}} \in \mathbb{R}^{(t_{\text{n}}+1) \times (t_{\text{n}}+1)}$ is defined as the output leaky matrix:
\begin{equation}
\begin{aligned}
\label{formula.8}
& E_{\mathrm{out}} =
    \begin{bmatrix}
        \epsilon_{\mathrm{out}}^0 & \epsilon_{\mathrm{out}}^1 & \hdots & \epsilon_{\mathrm{out}}^{\mathrm{t_{\text{n}}}}
    \end{bmatrix}, \\
& O = 
    \begin{bmatrix}
     o^0 & o^1 & \hdots & o^{\mathrm t_{\text{n}}}
    \end{bmatrix}, \\
& \bm{\mathcal{L}}_{\mathrm{out}} =
        \begin{bmatrix}
        0 & V_{\mathrm{th}} & V_{\mathrm{th}} \mathcal{L}(1) & \hdots & V_{\mathrm{th}} \mathcal{L}(t_{\mathrm{n}}-1) \\
        0 & 0 & V_{\mathrm{th}} & \hdots & V_{\mathrm{th}} \mathcal{L}(t_{\mathrm{n}}-2) \\
        0 & 0 & 0 & \hdots & V_{\mathrm{th}} \mathcal{L}(t_{\mathrm{n}}-3) \\
        \vdots & \vdots & \vdots & \ddots & \vdots \\
        0 & 0 & 0 & \hdots & 0
        \end{bmatrix}.
\end{aligned}
\end{equation}
Therefore, the firing process dynamics are represented. The neuron dynamics consisting of input stimulation, accumulation, and firing process are expressed as:
\begin{equation}
\label{formula.accurate_dyn}
\begin{aligned}
u^{\mathrm{t}} 
    & = \epsilon_{\mathrm{in}}^{t} - \epsilon_{\mathrm{out}}^{\mathrm{t}} \\
    & = \sum^{t}_{i=0}wx^{\mathrm{i}} \mathcal{L}(t-i) - \sum^{t-1}_{i=0}o^{\mathrm{i}}V_{\mathrm{th}}\mathcal{L}(t-i-1),
\end{aligned}
\end{equation}
with the matrix format as:
\begin{equation}
\label{formula.11}
U = E_{\mathrm{in}} - E_{\mathrm{out}} = E_{\mathrm{in}} - O \bm{\mathcal{L}_{\mathrm{out}}}.
\end{equation}

In \eqref{formula.accurate_dyn}, the output spikes $o$ are the only part that needs to be solved. Given the accurate membrane potential $U$ and the Heaviside function $g(\cdot)$, we have the following identity:
\begin{equation}
\label{formula.identity}
    O = g(U) = g(E_{\mathrm{in}} - O \bm{\mathcal{L}_{\mathrm{out}}}),
\end{equation}
where only $O$ is the unknown variable. To solve \eqref{formula.11}, we propose the following proposition:

\begin{proposition}
\label{proposition.os}
    Given a LIF neuron with stimulus, accumulation, and firing dynamics \eqref{formula.11}, the inequality
    \begin{equation}
        g(E_{\mathrm{in}}-I_{\mathrm{1}}{\bm{\mathcal{L}_{\mathrm{out}}}}) \leq O \leq g(E_{\mathrm{in}}) 
    \end{equation}
    always holds where $O \in \mathbb{R}^{1 \times (t_{\mathrm{n}}+1)}$ is the output spike matrix, $I_{\mathrm{1}} \in \mathbb{R}^{1 \times (t_{\mathrm{n}}+1)}$ is the all-ones matrix and ${\bm{\mathcal{L}_{\mathrm{out}}}}$ is defined in \eqref{formula.6}.
\end{proposition}
\textit{Proof:}
We use \eqref{formula.identity} to prove Proposition~\ref{proposition.os}. 
Because \( O = g(U) \), we have $\min(g(U)) \leq O \leq \max(g(U))$.
This implies $g(\min(U)) \leq O \leq g(\max(U))$.
Given \( U = E_{\mathrm{in}} - O \bm{\mathcal{L}_{\mathrm{out}}} \), we can write:
\begin{align}
g(\min(E_{\mathrm{in}} - O \bm{\mathcal{L}_{\mathrm{out}}})) \leq O \leq g(\max(E_{\mathrm{in}} - O \bm{\mathcal{L}_{\mathrm{out}}})).
\end{align}
Then, we have:
\begin{align}
g(E_{\mathrm{in}} - \max(O) \bm{\mathcal{L}_{\mathrm{out}}}) \leq O \leq g(E_{\mathrm{in}} - \min(O) \bm{\mathcal{L}_{\mathrm{out}}}).
\end{align}
Since $O$ represents the binary output spikes matrix, it follows that $I_{\mathrm{0}} \leq O \leq I_{\mathrm{1}}$.
Thus, we obtain:
\begin{align}
g(E_{\mathrm{in}} - I_{\mathrm{1}} \bm{\mathcal{L}_{\mathrm{out}}}) \leq O \leq g(E_{\mathrm{in}} - I_{\mathrm{0}} \bm{\mathcal{L}_{\mathrm{out}}}).
\end{align}
which finally simplifies as Proposition~\ref{proposition.os}. \hfill \( \square \)

    Thus, we use Proposition \ref{proposition.os} to estimate $O$ in \eqref{formula.11} as $U =E_{\mathrm{in}} - {\hat O}\bm{\mathcal{L}_{\mathrm{out}}}$,where $\hat O$ is the estimated output spikes. To ensure the sparsity of the output spikes (i.e., to minimize $O$), it is essential to consider the case of minimal $U$. Therefore, the membrane potential $U$ is estimated as:
    \begin{equation}
    \label{formula_non_iterative_lif_all_matrix}
    \begin{aligned}
        U & = E_{\mathrm{in}}-{\max(O)}{\bm{\mathcal{L}_{\mathrm{out}}}} = E_{\mathrm{in}}-g(E_{\mathrm{in}}){\bm{\mathcal{L}_{\mathrm{out}}}}\\
        O & = g(U).
    \end{aligned}
    \end{equation}

Therefore, the neuron dynamics of the NiLIF model with the soft reset mechanism can be represented as:
\begin{equation}
\label{formula_non_iterative_lif_all}
    \begin{aligned}
     u^{\mathrm t} & = \sum^{t}_{i=0}wx^{\mathrm{i}}\mathcal{L}(t-i) - \\
        & \sum^{t-1}_{i=0}g(\sum^{i}_{k=0}wx^{\mathrm{k}}\mathcal{L}(i-k))V_{\mathrm{th}}\mathcal{L}(t-1-i), \\
         o^{\mathrm t} & = g(u^{\mathrm t}).
    \end{aligned}
\end{equation}

We aim to preserve the input gradient values without introducing significant neuron changes. Therefore, we utilize the derivative as follows during backpropagation:
\begin{equation}
\label{eq:derivation_nilif}
    \frac{\partial o}{\partial u}=\left\{
    \begin{array}{ll}
         1, & \text{ if } 0 < u < 1,\\
         0, & \text{ else}.
    \end{array}
    \right.
\end{equation}

The pseudo-code of the NiLIF model is shown in Algorithm \ref{algo.neuron}. We present the derivations of the dynamics in this section.

\begin{algorithm}
    \caption{Pseudocode for the NiLIF model}
    \label{algo.neuron}
    \begin{algorithmic}
     \State \textbf{Input} Time steps $t$, threshold $v$, decay constant $\tau$, input $x$.
        \Function{L\_in\_matrix}{$t$, $\tau$}
            \State $e \gets \text{zeros}((t, t))$
            \For{$i,j \in \text{range}(t)$}
                \If{$j>i$}
                    \State $e[i][j] \gets \exp(-(((j-i)) / \tau))$
                \EndIf
                \State \textbf{if } $i=j$ \textbf{ then } $e[i][j]=1$ \textbf{ end if}
            \EndFor
            \State \Return $e$
        \EndFunction

        \Function{L\_out\_matrix}{$e$, $v$}
            \State $s \gets \text{zeros}(e)$
            \For{$i,j \in \text{range}(t)$}
                    \State \textbf{if } $j > i$ \textbf{ then } $s[i][j]=e[i][j-1]v$ \textbf{ end if}
            \EndFor
            \State \Return $s$
        \EndFunction
        
        \Procedure{Ni\_LIF}{}
            \Function{\_\_init\_\_}{$t$, $v$}
                \State $e \gets \Call{L\_in\_matrix}{t}$, $s \gets \Call{L\_out\_matrix}{e, v}$
            \EndFunction
        
            \Function{forward}{$x$}
                \State $u \gets xe$, $o \gets (u > v)s$, $u \gets u - o$, $o \gets (u > v)$
                \State \Return $o$
            \EndFunction
        \EndProcedure
        
    \end{algorithmic}
\end{algorithm}

\subsubsection{Comparison: The Iterative LIF and NiLIF neurons}
\label{Methods.lif.comparison}

The flow diagrams for the two neurons are shown in Figure \ref{fig_neuron_comparison}. The Iterative LIF neuron operates in a recurrent manner, relying on the output from the preceding time step for its next computation. Conversely, the Non-iterative LIF model simultaneously processes data from all time steps, requiring only a few matrix operations to determine the membrane potential and output spikes for all time steps. The non-loop characteristic of the Non-iterative LIF neuron could avoid the gradient issues caused by long-term dependencies. We assume the loss is $L \in \mathbb{R}^{1 \times (t_{\text{n}}+1)}$, which is equal to the summation of the loss $l^{\mathrm t} \in \mathbb{R}$ of each time step $t$. 
The gradient derivation for both LIF neuron models during backpropagation is shown as:
\begin{equation}
    \frac{\partial L}{\partial w} 
    = \sum^{t_{\mathrm n}}_{t=0} \frac{\partial l_{\mathrm t}}{\partial w} = \sum^{t_{\mathrm n}}_{t=0} \frac{\partial l_{\mathrm t}}{\partial o^{\mathrm t}} \frac{\partial o^{\mathrm t}}{\partial u^{\mathrm t}} \frac{\partial u^{\mathrm t}}{\partial w}.
\end{equation}
Therefore, we introduce two Propositions regarding $\frac{\partial u^{\mathrm t}}{\partial w}$ in two LIF neuron models respectively to compare the gradient issues. Proposition \ref{proposition:gradient_I-LIF} shows the gradient equation of the Iterative LIF neuron, and Proposition \ref{proposition:gradient_Ni-LIF} introduces the gradient equation of the Non-iterative LIF neuron.
\begin{proposition}
    \label{proposition:gradient_I-LIF}
     Given an Iterative LIF neuron with the dynamics in \eqref{formula_iterative_lif}, the limit condition
     \begin{equation}
         \forall u^{\mathrm t}_{\text{I}} \text{, } \lim\limits_{t \to \infty} \frac{\partial u^{\mathrm t}_\text{I}}{\partial w} = 0
\end{equation} always holds, where $u^{\mathrm t}_{\text{I}}$ is the membrane potential of the Iterative LIF neuron at the time step $t$.
\end{proposition}

\textit{Proof: }
    In the Iterative LIF neuron model, $\frac{\partial u^{\mathrm t_{\text{n}}}_\text{I}}{\partial w}$ is derived in \eqref{eq:gradient_I-LIF}, 
    which is composed of a summation of two parts of accumulated gradient multiplication $\prod^{t_{\text{n}}-1}_{i=0}\frac{\partial u^{\mathrm i+1}_{\text{I}}}{\partial u^{\mathrm i}_{\text{I}}}$. According to \eqref{formula_iterative_lif}, $\frac{\partial u^{\mathrm i+1}_{\text{I}}}{\partial u^{\mathrm i}_{\text{I}}}$ is equal to the decay rate $\lambda \in (0, 1)$ which is less than 1. Therefore, $\lim\limits_{t \to \infty} \prod^{t-1}_{i=0} \frac{\partial u^{\mathrm i+1}_{\text{I}}}{\partial u^{\mathrm i}_{\text{I}}} = \lim\limits_{t \to \infty} \lambda^t = 0$, which causes $\lim\limits_{t \to \infty} \frac{\partial u^{\text{t}}_{\text{I}}}{\partial w} = 0$.
    \hfill \( \square \)

\begin{proposition}
\label{proposition:gradient_Ni-LIF}
     Given a NiLIF neuron with the dynamics in \eqref{formula_non_iterative_lif_all}, the limit condition
     \begin{equation}
         \exists u^{\mathrm t}_{\text{Ni}} \text{, s.t.} \lim\limits_{t \to \infty} \frac{\partial u^{\mathrm t}_{\text{Ni}}}{\partial w} \neq 0
     \end{equation}
     always holds where $u^{\mathrm t}_{\text{Ni}}$ is the membrane potential of the NiLIF neuron at the time step $t$.
\end{proposition}

\textit{Proof: }
    We construct a specific example to prove $\exists u^{\mathrm t_{\text{n}}}_{\text{Ni}} \text{, s.t.} \lim\limits_{t_{\text{n}} \to \infty} \frac{\partial u^{\mathrm t_{\text{n}}}_{\text{Ni}}}{\partial w} \neq 0$.
    We use the symbol $f(n)$ to represent the term $\sum^{n}_{i=0}wx^{\mathrm{i}}\mathcal{L}(t-i)$ in \eqref{formula_non_iterative_lif_all}. Therefore, the membrane potential $u_{\mathrm{Ni}}^{\mathrm t}$ can be represented as:
    \begin{equation}
        \begin{aligned}
            u^{\mathrm t}_{\mathrm{Ni}} = f(t) - \sum^{t-1}_{i=0} g(f(i)) V_{\mathrm{th}} \mathcal{L}(t-1-i).
        \end{aligned}
    \end{equation}
    Therefore, $\frac{\partial u^{\mathrm t}_{\mathrm{Ni}}}{\partial w}$ is derived as:
    \begin{equation}
    \label{eq:derivation_1}
        \begin{aligned}
            \frac{\partial u^{\mathrm t}_{\mathrm{Ni}}}{\partial w} = \frac{\partial f(t)}{\partial w} - \sum^{t-1}_{i=0} \frac{\partial g(f(i))}{\partial f(i)} \frac{\partial f(i)}{\partial w} V_{\mathrm{th}} \mathcal{L}(t-1-i).
        \end{aligned}
    \end{equation}
    There exists $w>0$, $x^{\mathrm{t}}>0$, $\forall i<t, x^{\mathrm{i}}<0$ such that $\forall i<t, f(i)<0$ and $f(t) \in (0, 1)$. In this case, $\frac{\partial g(f(i))}{\partial f(i)}$ is equal to 0 based on \eqref{eq:derivation_nilif}. \eqref{eq:derivation_1} is derived as:
    \begin{equation}
        \begin{aligned}
            \frac{\partial u^{\mathrm t}_{\mathrm{Ni}}}{\partial w} = \frac{\partial f(t)}{\partial w} = \sum^{t}_{i=0}x^{\mathrm{i}}\mathcal{L}(t-i).
        \end{aligned}
    \end{equation}
    In the case of $\sum^{t}_{i=0}wx^{\mathrm{i}}\mathcal{L}(t-i)$ is in $(0, 1)$ and $w>0$, $\sum^{t}_{i=0}x^{\mathrm{i}}\mathcal{L}(t-i)$ is not 0.
    Therefore, we prove that $\exists u^{\mathrm t_{\text{n}}}_{\text{Ni}} \text{, s.t.} \lim\limits_{t_{\text{n}} \to \infty} \frac{\partial u^{\mathrm t_{\text{n}}}_{\text{Ni}}}{\partial w} \neq 0$.
    \hfill \( \square \)

Proposition \ref{proposition:gradient_I-LIF} and \ref{proposition:gradient_Ni-LIF} show that when the number of time steps is large, the iterative LIF neuron has the vanishing gradient problem, but the NiLIF neuron does not.
As well known in \cite{hochreiter2001gradient}, the presence of a continuously accumulated gradient multiplication part, $\prod^{t_{\text{n}}-1}_{i}\frac{\partial u^{\mathrm i+1}_{\text{I}}}{\partial u^{\mathrm i}_{\text{I}}}$, can cause gradient vanishing issues during training. Conversely, the gradient equations of the proposed NiLIF only contain the summation term, which can avoid the gradient issue caused by accumulated multiplication. The detailed comparison of gradient problems is in the appendix.

\begin{figure*}
{\small
    \begin{equation}
        \begin{aligned}
        \label{eq:gradient_I-LIF}
    \frac{\partial u^{\text{t}}_{\text{I}}}{\partial w} 
    &= \frac{\partial u^{\text{t}}_{\text{I}} }{\partial u^{\mathrm {t-1}}_{\text{I}}} \frac{\partial u^{\mathrm {t-1}}_{\text{I}} }{\partial w} + \frac{\partial u^{\text{t}}_{\text{I}}}{\partial o^{\mathrm {t-1}}} \frac{\partial o^{\mathrm {t-1}}}{\partial w} + x^{\mathrm {t-1}}  \\
    &= \frac{\partial u^{\text{t}}_{\text{I}}}{\partial u^{\mathrm {t-1}}_{\text{I}}} \left[ \frac{\partial u^{\mathrm {t-1}}_{\text{I}}}{\partial u^{\mathrm t-2}_{\text{I}}} \frac{\partial u^{\mathrm t-2}_{\text{I}}}{\partial w} + \frac{\partial u^{\mathrm t-2}_{\text{I}}}{\partial o^{\mathrm t-2}} \frac{\partial o^{\mathrm t-2}}{\partial w} + x^{\mathrm t-2}\right]  + \frac{\partial u^{\mathrm t}_{\text{I}}}{\partial o^{\mathrm {t-1}}} \frac{\partial o^{\mathrm {t-1}}}{\partial w} + x^{\mathrm {t-1}}  \\
    &= \frac{\partial u^{\text{t}}_{\text{I}}}{\partial u^{\mathrm {t-1}}_{\text{I}}} \left[ \frac{\partial u^{\mathrm {t-1}}_{\text{I}}}{\partial u^{\mathrm t-2}_{\text{I}}} \left[ \frac{\partial u^{\mathrm t-2}_{\text{I}}}{\partial u^{\mathrm t-3}_{\text{I}}} \left[ \hdots \left[ \frac{\partial u^2_{\text{I}}}{\partial u^1_{\text{I}}} \left[ \frac{\partial u^1_{\text{I}}}{\partial u^0_{\text{I}}} \frac{\partial u^0_{\text{I}}}{\partial w} \!+\! \frac{\partial u^1_{\text{I}}}{\partial o^0} \frac{\partial o^0}{\partial w} \!+\! x^0 \right] \!+\! \frac{\partial u^2_{\text{I}}}{\partial o^1} \frac{\partial o^1}{\partial w} \!+\! x^1 \right] \!+\! \hdots \right] \!+\!  \frac{\partial u^{\mathrm t-2}_{\text{I}}}{\partial o^{\mathrm t-3}} \frac{\partial o^{\mathrm t-3}}{\partial w} \!+\! x^{t-3} \right] \!+\! \frac{\partial u^{\mathrm {t-1}}_{\text{I}}}{\partial o^{\mathrm t-2}} \frac{\partial o^{\mathrm t-2}}{\partial w} \right.  \\
    & \quad+ x^{\mathrm t-2} \bigg]  + \frac{\partial u^{\text{t}}_{\text{I}}}{\partial o^{\mathrm {t-1}}} \frac{\partial o^{\mathrm {t-1}}}{\partial w} + x^{\mathrm {t-1}}  \\
    &= \sum^{t_{\text{n}}}_{i=1} \left[ \prod^{t_{n}-1}_{k=i} \frac{\partial u^{\mathrm k+1}_{\text{I}}}{\partial u^{\mathrm k}_{\text{I}}} \right]x^{i-1} +  \sum^{t_{\text{n}}}_{i=2} \left[ \prod^{t_{n}-1}_{k=i} \frac{\partial u^{\mathrm k+1}_{\text{I}}}{\partial u^{\mathrm k}_{\text{I}}} \right]\frac{\partial u^{\mathrm i}_{\text{I}}}{\partial o^{\mathrm i-1}}\frac{\partial o^{\mathrm i-1}}{\partial w}  = \sum^{t_{\text{n}}}_{i=2} \underbrace{\left[\prod^{t_{n}-1}_{k=i} \frac{\partial u^{\mathrm k+1}_{\text{I}}}{\partial u^{\mathrm k}_{\text{I}}} \right]}_{\substack{\text{Accumulating} \\ \text{gradient mult.}}} [ x^{i-1}+ \frac{\partial u^{\mathrm i}_{\text{I}}}{\partial o^{\mathrm i-1}} \underbrace{\frac{\partial o^{\mathrm i-1}}{\partial w}}_{\substack{\text{Accumulating} \\ \text{gradient mult.}}} ] + \underbrace{\prod^{t_{n}-1}_{i=1} \frac{\partial u^{\mathrm i+1}_{\text{I}}}{\partial u^{\mathrm i}_{\text{I}}}}_{\substack{\text{Accumulating} \\ \text{gradient mult.}}} x^0.
    \end{aligned}
    \end{equation}
    }
    \vspace{-1cm}
    \end{figure*}
\begin{figure}
	\centering
		\includegraphics[width=\linewidth]{ 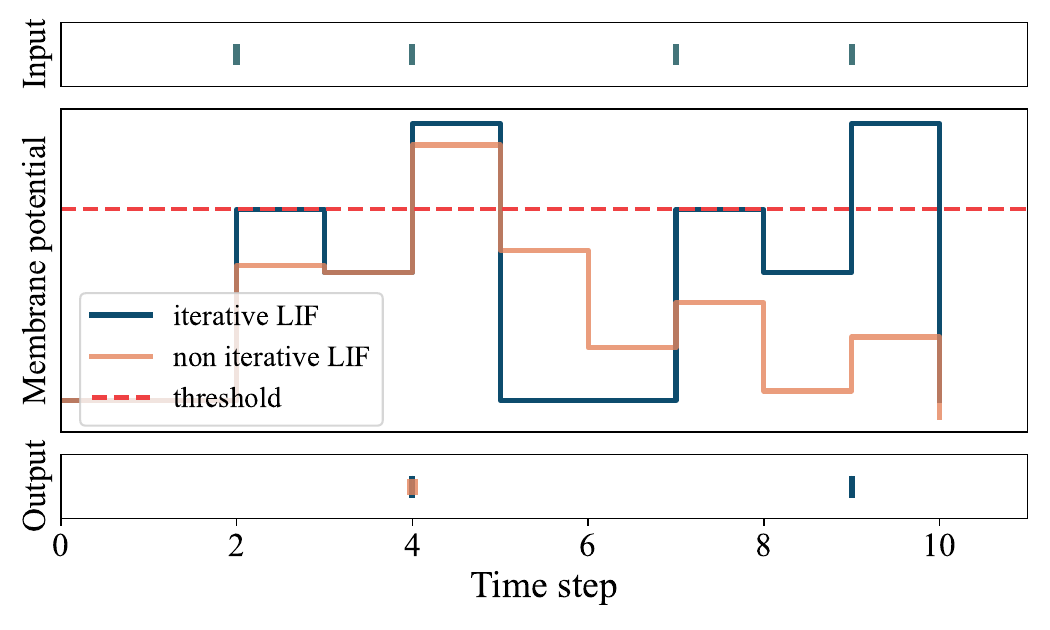}
	  \caption{The illustrative comparison between the Iterative LIF and NiLIF models. Both models utilize the same input spike train as shown in the uppermost figure, however, with different outputs. From the third figure, both models produce an output spike at time step $4$. Notably, the NiLIF model does not generate a second spike at time step $9$ due to higher sparsity.}\label{fig_neuron_comparison_spikes}
\end{figure}

In our NiLIF model, the final membrane potential $U$ is decided by the approximate term $g(E_{\mathrm{in}})$ and takes the minimal border of the actual situation. $E_{\mathrm{in}}$ overestimates the output spikes, resulting in $g(E_{\mathrm{in}})$ being greater than it should be, yielding a reduced final membrane potential $U$ and fewer output spikes in $O$. Therefore, NiLIF exhibits greater sparsity than the Iterative LIF model, as shown in Fig.~\ref{fig_neuron_comparison_spikes}. Given the same input spike train, the NiLIF neuron produces fewer spikes than its iterative counterpart. The sparsity brings more energy efficiency during execution.

\subsection{Attention model}
\label{Methods.attention}

In this section, we present the proposed attention models. The fundamental goal of the attention mechanism is to employ neural networks to compute the attention score, which is applied across the entire feature map, assigning weights to the features and extracting useful ones. In the context of EEG signals, the original input data is shaped as $\mathbb{R}^{B \times C \times D}$, where $B \in \mathbb{N}$ is the batch size, $C \in \mathbb{N}$ represents the channel size and $D \in \mathbb{N}$ is the length of data for each channel. We segment the data into timepieces, resulting in a new size of $\mathbb{R}^{B \times C \times S \times T}$, where $S$ is the number of timepieces and $T \in \mathbb{N}$ is the number of time steps in each segment. It is important to note that the multiplication result of $S$ and $T$ should equal $D$. Given that EEG data typically presents as long temporal sequences, it is important to determine which timepieces are crucial for classification. Thus, our attention model places particular emphasis on the dimension $S$. We introduce two distinct attention mechanisms: Sequence attention (Seq-attention), described in Section~\ref{Methods.attention.seq_attention}, and Channel Sequence attention (ChanSeq-attention), detailed in Section~\ref{Methods.attention.chanseq_attention}. Section~\ref{Methods.attention.global} presents Global-attention, a special case of ChanSeq-attention. 

\begin{figure}[tp]
	\centering
		\includegraphics[width=.6\linewidth]{ 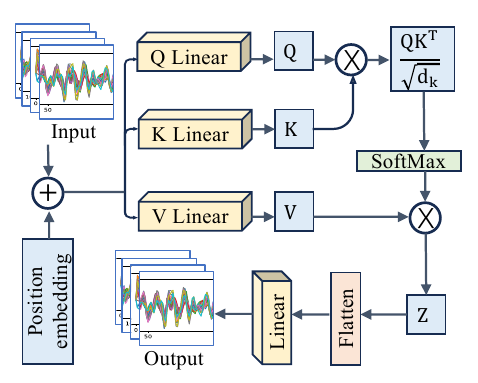}
      \vspace{-0.4cm}
	  \caption{Illustration of linear attention model. The linear attention model integrates position embedding with the input data. This model employs three linear layers to obtain the Query, Key, and Value matrices. The attention score is computed through the multiplication of the query and key matrices, then undergoing normalization through a Softmax function. To produce the final enhanced output, an additional linear layer is utilized in the final step.}
   \vspace{-0.5cm}
   \label{fig_linear_attention}
\end{figure}

\begin{figure}[tp]
	\centering
		\includegraphics[width=.6\linewidth]{ 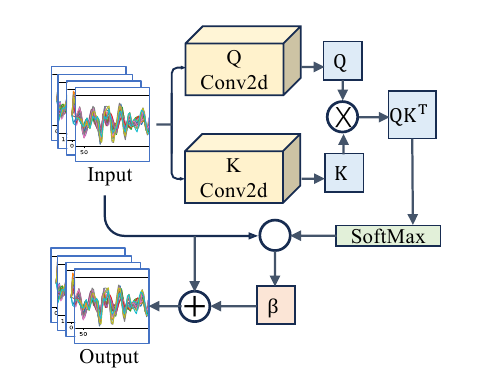}
      \vspace{-0.4cm}
	  \caption{Illustration of convolutional attention model. Two convolutional layers are employed to get the Query and Key matrices in the convolutional attention model. The attention score is computed through matrix multiplication and subsequently normalized by the Softmax function. A matrix operation then integrates the attention score with the input data. Depending on the model variant, this can manifest as matrix multiplication in the Seq-attention model (refer to Section~\ref{Methods.attention.seq_attention}) and the ChanSeq-attention model (detailed in Section~\ref{Methods.attention.chanseq_attention}), or as an element-wise product in the Global-attention model (see Section~\ref{Methods.attention.global}). In the final step, a trainable parameter, denoted as $\beta$, is introduced to balance the original and refined features.}
   \vspace{-0.4cm}
   \label{fig_conv_attention}
\end{figure}

Contrary to the sequential attention models like those in \cite{yao2023attention}, our intention is to utilize a single model to capture the attention score simultaneously. We introduce two model architectures for each attention mechanism: the linear architecture and the convolutional architecture. Fig.~\ref{fig_linear_attention} illustrates how the architecture of the linear attention differs from the attention model described in Fig.~\ref{fig_conv_attention}. The distinction between these architectures lies in their methods for attention score computation: the linear architecture incorporates fully connected layers, whereas the convolutional architecture employs convolutional layers. Notably, both Seq-attention and ChanSeq-attention have linear and convolutional versions.

\subsubsection{Seq-attention mechanism}
\label{Methods.attention.seq_attention}

The Sequence attention mechanism mainly focuses on identifying which timepieces require attention. The linear Seq-attention (Linear-Seq-attention) model are given as follows after reshaping the input data into dimension $\mathbb{R}^{B \times S \times (C \times T)}$:
\begin{equation}
\begin{aligned}
\label{formula_linear_seq_attention}
    & q = Q_{\mathrm{fc}}(x+p), \; Q_{\mathrm{fc}}: \mathbb{R}^{B \times S \times (C \times T)} \rightarrow \mathbb{R}^{B \times d_1 \times S \times d_2},  \\
    & k = K_{\mathrm{fc}}(x+p), \; K_{\mathrm{fc}}: \mathbb{R}^{B \times S \times (C \times T)} \rightarrow \mathbb{R}^{B \times d_1 \times S \times d_2},  \\
    & v = V_{\mathrm{fc}}(x+p), \; V_{\mathrm{fc}}: \mathbb{R}^{B \times S \times (C \times T)} \rightarrow \mathbb{R}^{B \times d_1 \times S \times d_2} , \\
    & A = \mathrm{Softmax}\left(\frac{q k^{\top}}{\sqrt{d_k}}\right), \; A \in \mathbb{R}^{B \times d_1 \times S \times S}, \\
    & \hat x = A v, \; \hat x \in \mathbb{R}^{B \times d_1 \times S \times d_2}, \\
    & h_{\mathrm{linear-seq}}(x) = FC_{\mathrm{seq}}(\hat x), \; \\ 
    & FC_{\mathrm{seq}}: \mathbb{R}^{B \times d_1 \times S \times d_2} \rightarrow \mathbb{R}^{B \times C \times S \times T},
\end{aligned}
\end{equation}
where $Q_{\mathrm{fc}}$, $K_{\mathrm{fc}}$ and $V_{\mathrm{fc}}$ are three linear layers to generate the query matrix $q$, the key matrix $k$, and the value matrix $v$, respectively. In particular, these matrices adhere to dimensions $\mathbb{R}^{B \times d_1 \times S \times d_2}$, where $d_1$ and $d_2$ represent hyperparameters. $p$ is the position embedding \cite{vaswani2017attention}. Subsequently, the attention score $A$ is the matrix product derived by $qK^{\top}$, followed by normalization. By applying the Softmax function to $A$, weights are assigned to the values in $v$. Then, a fully connected layer produces the final output $h_{\text{linear-seq}}(x)$, which represents the input data enhanced with attention.

The Convolutional Seq-attention (Conv-Seq-attention) model are presented as follows:
\begin{equation}
\begin{aligned}
\label{formula_conv_seq_attention}
    & q = Q_{\mathrm{conv}}(x), \; Q_{\mathrm{conv}}: \mathbb{R}^{B \times S \times C \times T} \rightarrow \mathbb{R}^{B \times S \times (d \times C \times T)}, \\
    & k = K_{\mathrm{conv}}(x), \; K_{\mathrm{conv}}: \mathbb{R}^{B \times S \times C \times T} \rightarrow \mathbb{R}^{B \times S \times (d \times C \times T)}, \\
    & A = \mathrm{Softmax}(q k^{\top}), \; A \in \mathbb{R}^{B \times S \times S}, \\
    & \hat x = A x, \; \hat x \in \mathbb{R}^{B \times C \times S \times T}, \\
    & h_{\mathrm{conv-seq}} (x) = \alpha \hat x + x,
\end{aligned}
\end{equation}
where $Q_{\mathrm{conv}}$ and $K_{\mathrm{conv}}$ are two convolutional layers with reshape techniques. Similarly to the Linear-Seq-attention mechanism, they are designated to generate the query matrix $q$ and key matrix $k$, respectively. Within this model, $d$ serves as a hyperparameter. Unlike its linear counterpart, the Conv-Seq-attention model omits the computation of the value matrix and instead directly calculates the attention score $A$ by multiplying the matrix. Subsequent to the Softmax function, these weights are integrated with the input data via matrix multiplication. Finally, the Conv-Seq-attention introduces a trainable parameter $\alpha$ to modulate the balance between the attention-enhanced result and the original input data.

Both the Linear-Seq-attention and Conv-Seq-attention mechanisms yield attention scores of size $\mathbb{R}^{S \times S}$ in the last two dimensions. Then they utilize matrix multiplication to produce the final enhanced feature map. In this way, attention is exclusively directed toward different timepieces and their interaction without consideration of information from other dimensions.

\subsubsection{ChanSeq-attention mechanism}
\label{Methods.attention.chanseq_attention}

Typically, EEG signals are collected from multiple channels. For example, the OpenBMI dataset that we use in this work has 62 channels \cite{lee2019eeg}. Beyond directing attention to timepieces, it is also valuable to be concerned with which channels receive the most attention. Thus, we introduce an attention mechanism named the Channel Sequence attention mechanism, designed to determine when and where the features must be focused.

The equations for linear Channel Sequence attention (Linear-ChanSeq-attention) are as follows:
\begin{equation}
\begin{aligned}
\label{formula_linear_chanseq_attention}
    & q = Q_{\mathrm{fc}}(x+p), \; Q_{\mathrm{fc}}: \mathbb{R}^{B \times C \times S \times T} \rightarrow \mathbb{R}^{B \times C \times d_1 \times S \times d_2}, \\
    & k = K_{\mathrm{fc}}(x+p), \; K_{\mathrm{fc}}: \mathbb{R}^{B \times C \times S \times T} \rightarrow \mathbb{R}^{B \times C \times d_1 \times S \times d_2}, \\
    & v = V_{\mathrm{fc}}(x+p), \; V_{\mathrm{fc}}: \mathbb{R}^{B \times C \times S \times T} \rightarrow \mathbb{R}^{B \times C \times d_1 \times S \times d_2}, \\
    & A = \mathrm{Softmax}\left(\frac{q k^{\top}}{\sqrt{d_k}}\right), \; A \in \mathbb{R}^{B \times C \times d_1 \times S \times S}, \\
    & \hat x = A v, \; \hat x \in \mathbb{R}^{B \times C \times d_1 \times S \times d_2}, \\
    & h_{\mathrm{linear-chanseq}}(x) = FC_{\mathrm{chanseq}}(\hat x), \; \\ 
    & FC_{\mathrm{chanseq}}: \mathbb{R}^{B \times C \times d_1 \times S \times d_2} \rightarrow \mathbb{R}^{B \times C \times S \times T},
\end{aligned}
\end{equation}
where the meaning of the parameters is the same as in Linear-Seq-attention. However, it is worth noting that to implement attention across both channels and timepieces simultaneously, the dimensions of $Q_{\mathrm{fc}}$, $K_{\mathrm{fc}}$, and $V_{\mathrm{fc}}$ are adjusted to $\mathbb{R}^{B \times C \times d_1 \times S \times d_2}$ instead of $\mathbb{R}^{B \times d_1 \times S \times d_2}$. The new channel dimension $C$ is consistently maintained throughout the entirety of the model.

The equations of corresponding convolutional Channel Sequence attention (Conv-ChanSeq-attention) are as follows:
\begin{equation}
\begin{aligned}
\label{formula_conv_chanseq_attention}
    & q = Q_{\mathrm{conv}}(x), \; Q_{\mathrm{conv}}: \mathbb{R}^{B \times C \times S \times T} \rightarrow \mathbb{R}^{B \times C \times S \times (d \times T)}, \\
    & k = K_{\mathrm{conv}}(x), \; K_{\mathrm{conv}}: \mathbb{R}^{B \times C \times S \times T} \rightarrow \mathbb{R}^{B \times C \times S \times (d \times T)}, \\
    & A = \mathrm{Softmax}(q  k^{\top}), \; A \in \mathbb{R}^{B \times C \times S \times S}, \\
    & \hat x = A  x, \; \hat x \in \mathbb{R}^{B \times C \times S \times T}, \\
    & h_{\mathrm{conv-chanseq}} (x) = \alpha \hat x + x,
\end{aligned}
\end{equation}
where the parameters in this context hold the same representations to those in the Conv-Seq-attention. Mirroring the adjustments in Linear-ChanSeq-attention, $Q_{\mathrm{conv}}$ and $K_{\mathrm{conv}}$ have dimensions $\mathbb{R}^{B \times C \times S \times (d \times T)}$, which is different from the previous approach of merging the $C$ and $T$ dimensions. As a result, the attention score $A$ is characterized by a size of $\mathbb{R}^{B \times C \times S \times S}$, facilitating the attention across various channels while accounting for different timepieces.

\subsubsection{Global-attention mechanism}
\label{Methods.attention.global}

In this section, we go further from the ChanSeq-attention. Given input data with dimensions $\mathbb{R}^{B \times C \times S \times T}$, not only are channels and timepieces considered, but also the specific time steps are important. To address this, we introduce the Global-attention mechanism, which operates attention across all three dimensions: $C$, $S$, and $T$. It decides when, where, and which feature is essential.

The Global-attention can be viewed as a special case of the Conv-ChanSeq-attention when the dimensions of $S$ and $T$ are equal. The corresponding equations are presented below:
\begin{equation}
\begin{aligned}
\label{formula_conv_global_attention}
    & q = Q_{\mathrm{conv}}(x), \; Q_{\mathrm{conv}}: \mathbb{R}^{B \times C \times S \times T} \rightarrow \mathbb{R}^{B \times C \times S \times (d \times T)}, \\
    & k = K_{\mathrm{conv}}(x), \; K_{\mathrm{conv}}: \mathbb{R}^{B \times C \times S \times T} \rightarrow \mathbb{R}^{B \times C \times T \times (d \times S)}, \\
    & A = \mathrm{Softmax}(q k^{\top}), \; A \in \mathbb{R}^{B \times C \times S \times T}, \\
    & \hat x = A \odot x, \; \hat x \in \mathbb{R}^{B \times C \times S \times T},  \\
    & h_{\mathrm{conv-global}} (x) = \alpha \hat x + x.
\end{aligned}
\end{equation}
In this context, the parameters align with those defined in the Conv-ChanSeq-attention. In particular, when $S$ and $T$ have same dimension sizes, both $A$ and $x$ have the size of $\mathbb{R}^{B \times C \times S \times T}$. Contrary to the Conv-ChanSeq-attention, the Global-attention employs element-wise product instead of matrix multiplication when calculating enhanced feature map $\hat x$ in \eqref{formula_conv_global_attention}. Consequently, with $A$ having dimensions $\mathbb{R}^{B \times C \times S \times T}$, each time step in each timepiece of each channel receives a specific attention score to improve the feature map. This method distinguishes itself from other attention models by directly enhancing the data.

\subsection{Network Architecture}
\label{Methods.network}

\begin{figure*}
	\centering
    \subfloat[SNN architecture\label{fig_snn_structure}]{%
        \includegraphics[width=0.55\linewidth]{ 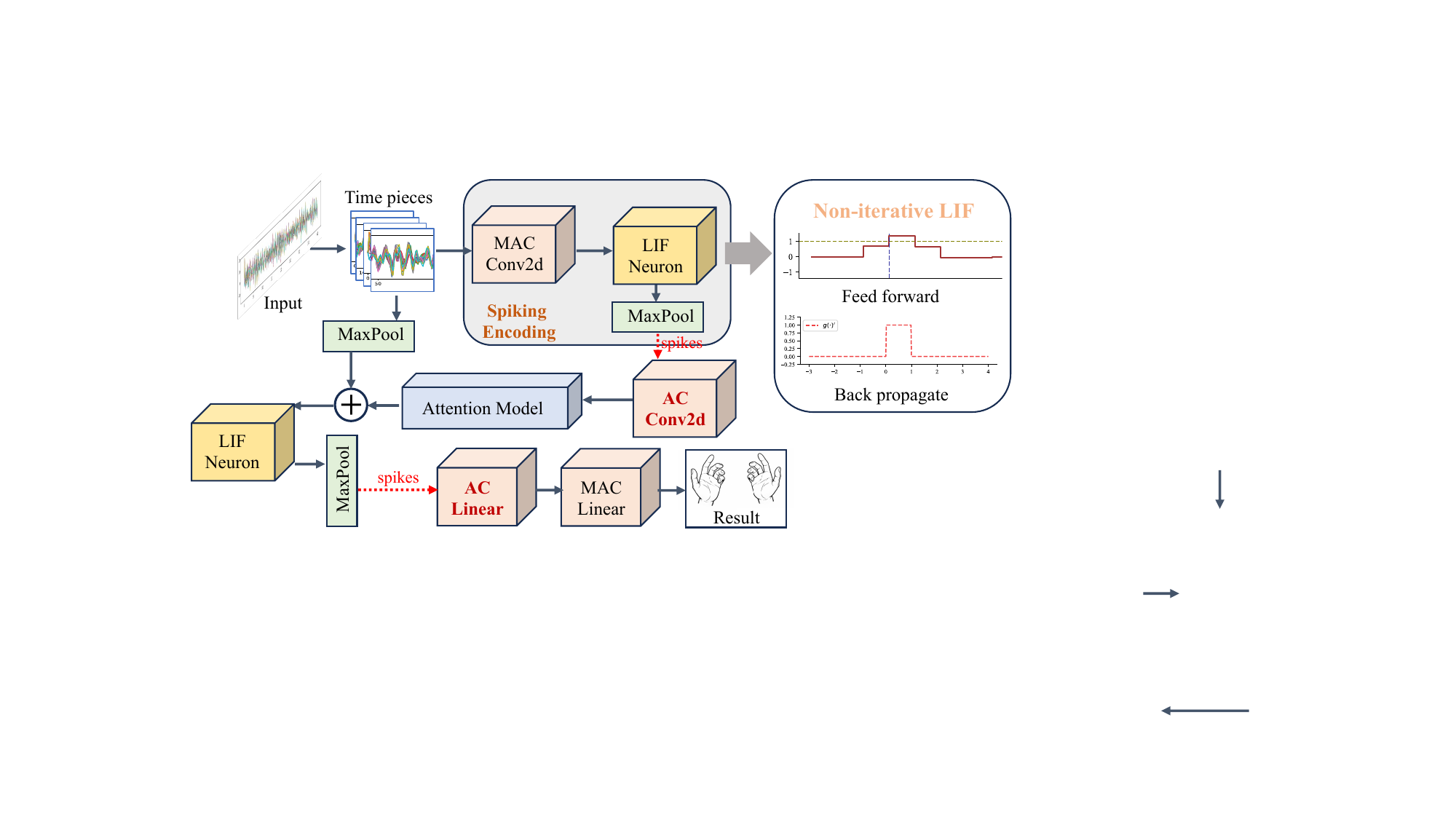}
    }
    \hspace{0.05\textwidth}
    \subfloat[CNN architecture\label{fig_cnn_structure}]{%
        \includegraphics[width=0.35\linewidth]{ 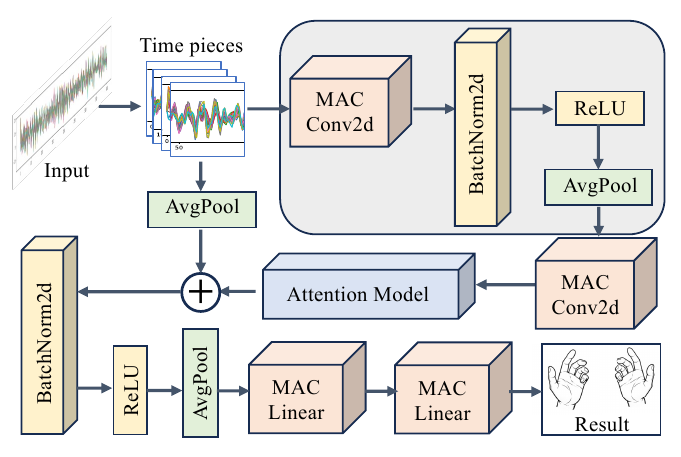}
    }
    \caption{
    Network architecture overview. (a) NiSNN-A architecture. The SNN has a residual block containing two spiking layers. Instead of traditional batch normalization layers and activation functions, the model employs LIF neurons to transform real-valued data streams into spikes. Notably, the proposed NiLIF neurons are utilized, leveraging the Heaviside function during the feed-forward phase and the surrogate derivation during backpropagation. Max pooling layers are adopted after spiking layers to maintain a binary data flow. After the second spiking layer, an attention mechanism is integrated to refine the feature maps. The final stage of the model consists of two fully connected layers for classification. (b) Attention CNN architecture. The overarching architecture of the CNN closely mirrors that of the SNN, encompassing a single residual block with two 2D convolutional layers. A batch normalization layer follows the convolutional layer and utilizes the ReLU function as its activation function. The network employs average pooling layers for down-sampling.
    }\label{fig_network_structure}
\end{figure*}

In this section, we present the network architectures shown in Fig.~\ref{fig_network_structure}. Fig.~\ref{fig_snn_structure} illustrates the architecture of NiSNN-A, which utilizes a two-layer residual spiking convolutional framework. The first spiking layer can be seen as a spiking encoder as proposed in \cite{liao2023convolutional}. The membrane potential batch normalization introduced in \cite{guo2023membrane} is also embedded in the NiLIF neurons. The NiLIF neuron yields a binary sequence as the output. To preserve the binary nature of the data stream, a max pooling layer is used to reduce dimensionality. After the second spiking layer, the proposed attention model is integrated. Finally, two linear layers without activation functions are employed to classify the output labels.

To compare and verify that the proposed attention mechanism can also be applied to the CNN network, we show the attention CNN counterpart to the NiSNN-A in Figure \ref{fig_cnn_structure}. The CNN architecture mirrors the SNNs to regulate extraneous variables, encompassing a two-layer convolutional residual framework. After each convolutional layer, a batch normalization layer is applied, followed by the ReLU activation function. The average pooling layer is then used to reduce the dimension.

In particular, since the SNN network has the characteristics of binary data flow, the input data for both the second spiking layer and the first linear layer consists of accumulator operations (AC). In contrast, all operations within the CNN are multiplicative and accumulate operations (MAC).

\section{Experiments}
\label{Experiments}

In this section, we outline the details of the experiment. Details about the dataset and its processing can be found in Section~\ref{Experiments.dataset}. The network configuration is elaborated in Section~\ref{Experiments.network}. Lastly, the approach to energy analysis is presented in Section~\ref{Experiments.energy}.

\subsection{Dataset}
\label{Experiments.dataset}

In this study, we utilized the BCIC IV 2a dataset \cite{tangermann2012review} and the OpenBMI dataset \cite{lee2019eeg} to evaluate the performance of the NiSNN-A.

The BCIC IV 2a dataset consists of EEG motor imagery recordings from 9 participants, with each participant having 288 trials across 20 channels. Each trial lasts 3 seconds and is recorded at a sampling frequency of 250 Hz. The OpenBMI dataset, which is larger in scale, includes EEG motor imagery signals from 54 participants, with each participant providing 400 trials across 62 channels. These trials are 4 seconds with a sampling frequency of 1000 Hz. The channels selected from OpenBMI include FC-5/3/1/2/4/6, C-5/3/1/z/2/4/6, and CP-5/3/1/z/2/4/6, whereas the selected channels for BCIC IV 2a are FC-3/1/z/2/4, C-5/3/1/z/2/4/6, CP-3/1/z/2/4, and P-1/z/2. The tasks of both datasets are to classify motor imagery EEG signals for left-hand and right-hand movements.

The experiments train the networks to identify the common features in EEG signals. We take the subject-independent approach as utilized in \cite{autthasan2021min2net}, which reserves data from one subject for testing and uses data from the remaining subjects for training. Specifically, for OpenBMI, the training data includes 53 subjects' recordings, totaling 21,200 trials, and the test data comprises 200 trials. For BCIC IV 2a, the training data includes 8 subjects' trials, totaling 2,304 trials, with the test dataset comprising 144 trials. For both datasets, each trial contains 400 time steps after downsampling. Therefore, this strategy ensures that the model is tested on unseen data, reinforcing its scalability and practical applicability.

Throughout the training phase, we cycle each subject as a test case. The overall performance is then calculated by averaging the accuracy from all cycles. Therefore, each model undergoes training and testing 54 times for OpenBMI and 9 times for BCIC IV 2a, respectively, to ascertain the final result and remove some uncertainty.

\subsection{Network Setups}
\label{Experiments.network}

As described in Section~\ref{Experiments.dataset}, the input data has a dimension size of $\mathbb{R}^{B \times C \times S \times T}$,  where $B$ denotes the batch size, $C$ is the channel size, $S$ indicates the number of timepieces, and $T$ is the number of time steps within each timepiece. In the experiment, the batch size is 64 and the channel size is 20. Each trial is segmented into 20 timepieces, rendering $S$ and $T$ as 20. Also, we set the threshold $V_{\mathrm{th}}$ in the NiLIF as 0.5.

The network parameters are shown in Table~\ref{tab_network_structure}. The first convolutional layer acts as a spiking encoder, only considering the information within each timepiece with a kernel size of $(1,5)$. The second convolutional layer plays the role of classifier, taking into account both intra-timepieces and inner-timepiece information with a kernel size of $(10, 10)$. All pooling layers employ the kernel size $(2, 2)$. The first linear layer reduces the flattened data dimension to 20, and the final linear layer reduces it further to 2, thereby having the final classification result. Within the attention layer, the hyperparameters $d_1$ and $d_2$ are set to 6 and 20, respectively, for all linear attention models. For convolutional attention models, the hyperparameter $d$ is set to 8. The models are trained for 20 epochs. We employ a well-trained CNN model as a pre-trained network for its corresponding SNN model to accelerate the training procedure. During the training process, we adopted the Adam optimizer with a learning rate of 0.001. The cross-entropy loss is utilized as the loss function:
\begin{equation}
    \mathrm{CE\_Loss} (y, p) = - \sum_{n}^{\mathrm n}y_\mathrm{n} \log(p_{\mathrm{n}}),
\end{equation}
where $y$ represents the label of the data and $p$ is the network's output.

\begin{table*}[!ht]
\caption{Architecture of networks during training. B is the batch size, C is channel size, S is the number of timepieces, and T is the number of time points in each timepiece.}\label{tab_network_structure}
\begin{tabularx}{\textwidth}{@{} *{6}{X} @{}}
\toprule
    Block & SNN layer & CNN layer & Filters & Size/padding & Output\\ 
    \midrule
    \multirow{6}*{Spike Encoding} & Input & Input & - & - & (B, C, S, T)\\
    ~ & Clone residual & Clone residual & - & - & (B, C, S, T)\\
    ~ & Conv2d & Conv2d & C & (1, 5)/same & (B, C, S, T)\\
    ~ & - & BatchNorm2d & C & - & (B, C, S, T)\\
    ~ & - & ReLU &  - & - & (B, C, S, T)\\
    ~ & LIF & - & - & - & (B, C, S, T)\\
    ~ & MaxPool2d & AvgPool2d & - & (2, 2) & (B, C, S/2, T/2) \\
    
    \midrule
    \multirow{9}*{Classifier} & \textbf{AC-Conv} & \textbf{MAC-Conv2d} & C & (10, 10)/same & (B, C, S/2, T/2) \\
    ~ & Attention & Attention & - & - & (B, C, S/2, T/2)\\
    ~ & Add residual & Add residual & - & - & (B, C, S/2, T/2)\\
    ~ & - & BatchNorm2d & C & - & (B, C, S/2, T/2)\\
    ~ & - & ReLU & - & - & (B, C, S/2, T/2)\\
    ~ & LIF & - & - & - & (B, C, S/2, T/2) \\
    ~ & MaxPool2d & AvgPool2d & - & (2, 2) & (B, C, S/4, T/4) \\
    ~ & Flatten & Flatten & - & - & (B, C$\times$S/4$\times$T/4) \\
    ~ & \textbf{AC-Linear} & \textbf{MAC-Linear} & - & - & (B, 20) \\
    ~ & Linear & Linear & - & - & (B, 2) \\
  
\bottomrule
\end{tabularx}
\end{table*}

\subsection{Energy analysis}
\label{Experiments.energy}

For analyzing the energy consumption of CNN and SNN models, we adopt the same energy analysis method in \cite{yao2023attention}, which calculates the network's floating point operations (FLOP).

The main operations within neural networks in this context can be categorically divided into three primary types: the convolutional layer, the linear layer, and matrix multiplication. The FLOPs associated with each of these operations can be described as:
\begin{equation}
\begin{aligned}
    & \mathrm{FLOPs}_{\mathrm{conv}}^{\mathrm n} = k^{\mathrm n}_{0}  k^{\mathrm n}_{1}  h^{\mathrm n}  w^{\mathrm n}  c^{\mathrm n}  c^{\mathrm n-1},  \\
    & \mathrm{FLOPs}_{\mathrm{fc}}^{\mathrm n} = i^{\mathrm n}  o^{\mathrm n},  \\
    & \mathrm{FLOPs}_{\mathrm{mm}} = m_1  n  m_2,
\end{aligned}
\end{equation}
where for the $n^{\mathrm{th}}$ convolutional layer, $k^{\mathrm n}_{0}$ and $k^{\mathrm n}_{1}$ denote the dimensions of the convolutional kernel, while $h^{\mathrm n}$ and $w^{\mathrm n}$ represent the dimensions of the output map. In addition, $c^{\mathrm n}$ and $c^{\mathrm n-1}$ specify the size of the input and output data channels, respectively. For the $n^{th}$ linear layer, $i^{\mathrm n}$ represents the input size, and $o^{\mathrm n}$ denotes the output size. Finally, for matrix multiplication involving two matrices of dimensions $[m_1, n]$ and $[n, m_2]$, the FLOPs are determined as the product of these three dimensions.

\begin{table*}[t!]
\caption{FLOPs of model executions. CNN models only have MAC operation, and SNN models have MAC, $\text{AC}_{\rm{conv}}$, and $\text{AC}_{\rm{fc}}$ operations. $x$ is 4,810,040, $y$ is 1,170,040, $a$ is 4,000,000 and $b$ is 10,000.}\label{table_flop}
\begin{threeparttable}
\begin{tabularx}{\textwidth}{@{} >{\hsize=0.65\hsize}X >{\hsize=0.84\hsize}X >{\hsize=0.1\hsize}X >{\hsize=0.24\hsize}X >{\hsize=0.22\hsize}X >{\hsize=0.18\hsize}X >{\hsize=0.18\hsize}X >{\hsize=0.22\hsize}X @{}}
\toprule
    Reference & Method & Type & MAC & $\text{Ratio}_{\mathrm{MAC}}$ & $\text{AC}_{\rm{conv}}$ & $\text{AC}_{\rm{fc}}$ & $\text{Ratio}_{\mathrm{AC}}$ \\
    \midrule

    Autthasan et al. \cite{autthasan2021min2net} & vanilla CNN & CNN & $x$ & baseline & - & - & -\\

    Huang et al. \cite{huang2021electroencephalogram} & vanilla CNN & CNN & $x$ & $- 0\%$ &  - & - & -\\

    Liu et al. \cite{liu2020parallel} & Sequence + Temporal attention & CNN & $x$+1,644,200 & $\uparrow 34.2\%$ &  - & - & - \\

    Luo et al. \cite{luo2023shallow} & Sequence + Temporal attention & CNN & $x$+2,469,600 & $\uparrow 51.3\%$ &  - & - & - \\

    Fan et al. \cite{fan2021bilinear}, Wang et al. \cite{wang2021residual} & Channel + Sequence + Temporal attention & CNN & $x$+45,200 & $\uparrow 0.9\%$ & - & - & - \\

    Zhang et al. \cite{zhang2023self} & Channel attention & CNN & $x$+180,000 & $\uparrow 3.7\%$ & - & - & -\\

    \midrule

    Wu et al. \cite{wu2018spatio} & vanilla SNN (Iterative LIF) & SNN & $y$-180,000 & $\downarrow 79.4\%$ & 0.807$a$  & 0.255$b$ & baseline \\

    Hu et al. \cite{hu2021advancing} & vanilla SNN & SNN & $y$ & $\downarrow 75.7\%$ & 0.352$a$ & 0.383$b$ & $\downarrow 56.36\%$\\

    Yao et al. \cite{yao2023attention} & Channel attention & SNN & $y$+3,600 & $\downarrow 75.6\%$ & 0.433$a$ & 0.356$b$ & $\downarrow 46.26\%$ \\

    Yao et al. \cite{yao2023attention} & Sequence attention & SNN & $y$+2,400 & $\downarrow 75.6\%$ & 0.407$a$ & 0.425$b$ & $\downarrow 49.51\%$ \\

    Yao et al. \cite{yao2023attention} & Temporal attention & SNN & $y$+2,400 & $\downarrow 75.6\%$ & 0.459$a$ & 0.331$b$ & $\downarrow 43.12\%$ \\

    Yao et al. \cite{yao2023attention} & Channel + Temporal attention & SNN & $y$+6,000 & $\downarrow 75.6\%$ & 0.457$a$ & 0.384$b$ &  $\downarrow 43.37\%$\\

    Yao et al. \cite{yao2023attention} & Channel + Sequence attention & SNN & $y$+6,000 & $\downarrow 75.6\%$ & 0.449$a$ & 0.362$b$ &  $\downarrow 44.30\%$\\

    Yao et al. \cite{yao2023attention} & Sequence + Temporal attention & SNN & $y$+4,800 & $\downarrow 75.6\%$ & 0.484$a$ & 0.367$b$ & $\downarrow 39.93\%$\\

    Yao et al. \cite{yao2023attention} & Channel + Sequence + Temporal attention & SNN & $y$+8,400 & $\downarrow 75.5\%$ & 0.500$a$ & 0.272$b$ & $\downarrow 38.09\%$\\

    \midrule

    Ours & Linear-Seq-attention & SNN & $y$+97,800 & $\downarrow 73.6\%$ & 0.397$a$ & 0.312$b$ & $\downarrow 50.80\%$\\

    Ours & Conv-Seq-attention & SNN & $y$+822,100 & $\downarrow 58.6\%$ & 0.358$a$ & 0.419$b$ &  $\downarrow 55.58\%$\\

    Ours & Linear-ChanSeq-attention & SNN & $y$+496,800 & $\downarrow 65.3\%$ & 0.382$a$ & 0.326$b$ & $\downarrow 52.61\%$ \\

    Ours & Conv-ChanSeq-attention & SNN & $y$+824,000 & $\downarrow 58.5\%$ & 0.345$a$ & 0.403$b$ &  $\downarrow \mathbf{57.17\%}$\\

    Ours & Global-attention & SNN & $y$+806,000 & $\downarrow 58.9\%$ & 0.348$a$ & 0.413$b$ &  $\downarrow 56.76\%$\\

\bottomrule
\end{tabularx}
\begin{tablenotes}
    \small
    \item[*] Only the first vanilla SNN uses the Iterative LIF neurons (specified already), otherwise all SNN models use the proposed Non-iterative LIF neuron model.
\end{tablenotes}
\end{threeparttable}
\end{table*}

In the CNN models, all network data consist of real-valued numbers, which makes all operations MAC. The energy analysis is divided into the FLOPs corresponding to the standard vanilla CNN model, denoted by $x$, and the FLOPs for the additional attention model. Given our consistent network architecture for all models, the value of $x$ remains constant, 4,810,040. 
In the SNN models, the FLOPs analysis is more complicated. As elaborated in \ref{Methods.network}, the inputs to the second convolutional and first linear layers are all binary. Consequently, these layers employ AC operations, which accumulate weight directly without involving multiplications. It is worth noting that the number of AC operations is contingent upon the spike rate, given that accumulation happens only when the input is 1. Thus, the spike rates of these two layers emerge as important parameters when quantifying the AC operations within SNN models. The rest of the SNN retains the use of MAC operations, mirroring the corresponding CNN models. 
Therefore, the FLOPs assessment for SNN models can be partitioned into three parts: AC of the binary layers, MAC of the rest of the SNN model, and the MAC of the additional attention model. The MAC of the rest of the SNN model could be divided into two parts: the MAC of LIF neurons and the MAC of the rest of the NiSNN model. We present the MAC of the rest SNN model with NiLIF neurons as $y$, which is a constant across all SNN models with NiLIF neurons, calculated as 1,170,040. The MAC of SNN with Iterative LIF neurons is less than $y$, which is calculated as $y-180,000$. It should be highlighted that while the FLOPs associated with Iterative LIF neurons are fewer than those of NiLIF neurons, the latter are implemented using matrix operations, whereas the former rely on loop structures, leading to increased execution time.
The original AC for the second convolutional layer is 4,000,000, denoted by $a$, while for the first linear layer, it is 10,000, represented by $b$. 

The final FLOPs comparison of models are shown in Table~\ref{table_flop}. Details on model selection are provided in Section~\ref{Result.snn}. We choose the vanilla CNN and SNN as baseline models for comparing MAC and AC operations, respectively. All spiking rates are calculated by the average of two dataset experiments. For each experiment, the spiking rates are calculated over all subjects. Table~\ref{table_flop} shows that the proposed model can reduce the energy consumption from both MAC and AC calculation.
After the FLOPs analysis, we could calculate the total energy cost. We adopt the same assumption as in \cite{yao2023attention} that the data for various operations are implemented as floating point 32 bits in 45nm technology, where the MAC energy is 4.6$pJ$ and the AC energy is $0.9pJ$. The detailed energy is listed in Table~\ref{tabel_snn_bcic} and \ref{tabel_snn_openbmi}.

\section{Result and Discussion}
\label{ResultAndDiscussion}

In this section, we describe the results of the proposed attention models which are delineated in Section~\ref{Result.snn}. Then, a discussion along with the result visualization is provided in Section~\ref{Result.discussion}.

\begin{table*}[t!]
\begin{threeparttable}
\caption{Accuracy and energy performance of Left H. vs Right H. subject-independent classification using the BCIC IV 2a dataset for SNNs with attention mechanisms.  }\label{tabel_snn_bcic}
\begin{tabularx}{\textwidth}{@{} >{\hsize=1.4\hsize}X >{\hsize=2.06\hsize}X >{\hsize=0.3\hsize}X >{\hsize=0.86\hsize}X >{\hsize=0.35\hsize}X >{\hsize=0.35\hsize}X >{\hsize=0.7\hsize}X @{}}
\toprule
    Reference & Method & Type & Accuracy & $\text{AC}_\text{conv}$ & $\text{AC}_\text{fc}$ & Energy ($\mu{J}$) \\
    \midrule

    Autthasan et al. \cite{autthasan2021min2net} & vanilla CNN & CNN & 0.6545 +/- 0.1053 & - & - & \textbf{23.569} \\

    Huang et al. \cite{huang2021electroencephalogram} & vanilla CNN & CNN & 0.6866 +/- 0.0885 & - & - &\textbf{23.569 }\\

    Liu et al. \cite{liu2020parallel} & Sequence + Temporal attention & CNN & 0.6641 +/- 0.1011 & - & - &31.626 \\

    Luo et al. \cite{luo2023shallow} & Sequence + Temporal attention & CNN & 0.6667 +/- 0.1191 & - & - &35.670 \\

    Fan et al. \cite{fan2021bilinear}, Wang et al. \cite{wang2021residual} & Channel + Sequence + Temporal attention & CNN & 0.6745 +/- 0.1144 & - & - &23.791 \\

    Zhang et al. \cite{zhang2023self} & Channel attention & CNN & \textbf{0.6875 +/- 0.0880} & - & - &24.451\\
    
    \midrule

    Wu et al. \cite{wu2018spatio} & vanilla SNN (\underline{with the Iterative LIF neuron}) & SNN &  0.5182 +/- 0.0607 & 0.8034 & 0.1050 & 7.744 \\
    

    Hu et al. \cite{hu2021advancing} & vanilla SNN & SNN & 0.6120 +/- 0.0842 & 0.3966 & 0.4311 & 7.165\\


    Yao et al. \cite{yao2023attention} & Channel attention & SNN & 0.5703 +/- 0.0614 & 0.4891 & 0.4159 & 7.515 \\

    Yao et al. \cite{yao2023attention} & Sequence attention & SNN & 0.5773 +/- 0.0478 & 0.4607 & 0.4423 & 7.407 \\

    Yao et al. \cite{yao2023attention} & Temporal attention & SNN & 0.6094 +/- 0.0794 & 0.5284 & 0.4246 & 7.651 \\

    Yao et al. \cite{yao2023attention} & Channel + Temporal attention & SNN & 0.5885 +/- 0.0961 & 0.5093 & 0.4278 & 7.600 \\

    Yao et al. \cite{yao2023attention} & Channel + Sequence attention & SNN & 0.5920 +/- 0.0746 & 0.4803 & 0.4256 & 7.496 \\

    Yao et al. \cite{yao2023attention} & Sequence + Temporal attention & SNN & 0.5521 +/- 0.0576 & 0.5687 & 0.4558 & 7.808 \\

    Yao et al. \cite{yao2023attention} & Channel + Sequence + Temporal attention & SNN & 0.5512 +/- 0.0817 & 0.5365 & 0.4218 & 7.710 \\
    
    \midrule

    Ours & Linear-Seq-attention & SNN & 0.6259 +/- 0.1020 & 0.4200 & 0.3218  & 7.727\\

    Ours & Conv-Seq-attention & SNN & 0.6493 +/- 0.1030  & 0.3778 & 0.4452 & 11.126 \\

    Ours & Linear-ChanSeq-attention & SNN & 0.6241 +/- 0.0862 & 0.4056 & 0.3380 & 9.631 \\

    Ours & Conv-ChanSeq-attention & SNN & \textbf{0.6580 +/- 0.1079} & 0.3701 & 0.4363 &11.107\\

    Ours & Global-attention & SNN & \textbf{0.6615 +/- 0.1219} & 0.3784 & 0.4433  &11.049 \\

\bottomrule
\end{tabularx}
\begin{tablenotes}
    \small
    \item[*] Only the first vanilla SNN uses the Iterative LIF neurons (specified already). Otherwise, all SNN models use the proposed Non-iterative LIF neuron model.
\end{tablenotes}
\end{threeparttable}
\end{table*}

\begin{table*}[t]
\begin{threeparttable}
\caption{Accuracy and energy performance of Left H. vs Right H. subject-independent classification using the OpenBMI dataset for SNNs with attention mechanisms.  }\label{tabel_snn_openbmi}
\begin{tabularx}{\textwidth}{@{} >{\hsize=1.4\hsize}X >{\hsize=2.06\hsize}X >{\hsize=0.3\hsize}X >{\hsize=0.86\hsize}X >{\hsize=0.35\hsize}X >{\hsize=0.35\hsize}X >{\hsize=0.7\hsize}X @{}}
\toprule
    Reference & Method & Type & Accuracy & $\text{AC}_\text{conv}$ & $\text{AC}_\text{fc}$ & Energy ($\mu{J}$) \\
    \midrule

    Autthasan et al. \cite{autthasan2021min2net} & vanilla CNN & CNN & 0.7368 +/-0.1320 & - & - & \textbf{23.569} \\

    Huang et al. \cite{huang2021electroencephalogram} & vanilla CNN & CNN & 0.7370 +/- 0.1314& - & - & \textbf{23.569 }\\

    Liu et al. \cite{liu2020parallel} & Sequence + Temporal attention & CNN & 0.7394 +/- 0.1280 & - & - & 31.626 \\

    Luo et al. \cite{luo2023shallow} & Sequence + Temporal attention & CNN & 0.7138 +/- 0.1411 & - & - & 35.670 \\

    Fan et al. \cite{fan2021bilinear}, Wang et al. \cite{wang2021residual} & Channel + Sequence + Temporal attention & CNN & 0.7386 +/- 0.1279 & - & - & 23.791 \\

    Zhang et al. \cite{zhang2023self} & Channel attention & CNN & \textbf{0.7405 +/- 0.1362} & - & - & 24.451\\

    \midrule

    Wu et al. \cite{wu2018spatio} & vanilla SNN (\underline{with the Iterative LIF neuron}) & SNN & 0.5036 +/- 0.0053 & 0.8114 & 0.4040 & 7.776\\
    

    Hu et al. \cite{hu2021advancing} & vanilla SNN & SNN & 0.6971 +/- 0.1329 & 0.3067 & 0.3340 &  6.840\\

    Yao et al. \cite{yao2023attention} & Channel attention & SNN & 0.6757 +/- 0.1279 & 0.3776 & 0.2966 & 7.113\\

    Yao et al. \cite{yao2023attention} & Sequence attention & SNN  & 0.6978 +/- 0.1257 & 0.3531 & 0.4083 & 7.020\\

    Yao et al. \cite{yao2023attention} & Temporal attention & SNN & 0.6975 +/- 0.1309 & 0.3891 & 0.2369 & 7.148\\

    Yao et al. \cite{yao2023attention} & Channel + Temporal attention & SNN & 0.6819 +/- 0.1265 & 0.4040 & 0.3395 & 7.220\\

    Yao et al. \cite{yao2023attention} & Channel + Sequence attention & SNN & 0.6848 +/- 0.1297 & 0.4180 & 0.2993 & 7.270\\

    Yao et al. \cite{yao2023attention} & Sequence + Temporal attention & SNN & 0.6934 +/- 0.1256 & 0.4002 & 0.2791 & 7.200\\

    Yao et al. \cite{yao2023attention} & Channel + Sequence + Temporal attention & SNN & 0.6774 +/- 0.1318 & 0.4626 & 0.1231 & 7.441\\

    \midrule

    Ours & Linear-Seq-attention & SNN & 0.6843 +/- 0.1254 & 0.3736 & 0.3016  & 7.560\\

    Ours & Conv-Seq-attention & SNN &\textbf{0.7073 +/- 0.1397} & 0.3379 & 0.3936 & 10.981\\

    Ours & Linear-ChanSeq-attention & SNN & 0.6845 +/- 0.1204 & 0.3587  & 0.3145 & 9.462\\

    Ours & Conv-ChanSeq-attention & SNN & \textbf{0.7083 +/- 0.1402} & 0.3201 & 0.3688 & 10.926\\

    Ours & Global-attention & SNN & 0.7051 +/- 0.1377 & 0.3183 & 0.3824 & 10.832\\

\bottomrule
\end{tabularx}
\begin{tablenotes}
    \small
    \item[*] Only the first vanilla SNN uses the Iterative LIF neurons (specified already), otherwise all SNN models use the proposed Non-iterative LIF neuron model.
\end{tablenotes}
\end{threeparttable}
\end{table*}

\subsection{Comparison with state-of-the-art methods}
\label{Result.snn}

Table~\ref{tabel_snn_bcic} and \ref{tabel_snn_openbmi} compare the performances of various attention mechanisms with CNNs and SNNs on two EEG datasets. We have chosen seven CNN models with attention mechanisms for EEG classification as our benchmark. Autthasam et al. \cite{autthasan2021min2net} have proposed a vanilla CNN model, employing a subject-independent approach for training and testing phases. Furthermore, Huang et al. \cite{huang2021electroencephalogram} incorporated residual blocks into the CNN model. Of the seven baseline models, five models employ attention mechanisms. It is worth noting that \cite{fan2021bilinear} and \cite{wang2021residual} employ a global attention mechanism that encompasses all three dimensions. However, their approach is characterized by extracting attention scores individually for each dimension after using the pooling methods to minimize unrelated dimensions. This stands differently from the methodology of our proposed Global-attention model.
A total of ten models were chosen as the SNN baselines. It is worth noting that in the context of EEG signal processing, the Iterative LIF model struggles with the long time steps, potentially leading to gradient issues. Therefore, we adopted the proposed NiLIF neuron in other baseline models. Concerning the attention mechanism, the baseline models are from the image attention SNN model developed for computer vision, as introduced in \cite{yao2023attention}. This approach composes various dimensional attention components sequentially to achieve the attention score, diverging from our proposed models that utilize a singular model. 

The accuracy metric was determined by the ratio of correctly classified samples to the total number of samples, as given by:
\begin{equation}
\text{Accuracy} = \frac{\text{TP}+\text{TN}}{\text{TP}+\text{TN}+\text{FP}+\text{FN}},
\end{equation}
where TP, TN, FP, and FN represent true positives, true negatives, false positives, and false negatives, respectively. For energy analysis, we adopted the unit $\mu J$ to quantify the energy consumption of the networks as discussed in Section~\ref{Experiments.energy}.

The experiment results show that proposed NiSNN-As can achieve best performance among SNN models, which are 0.6615 and 0.7 for the BCIC IV 2a dataset and OpenBMI dataset, respectively. This accuracy is comparable with the CNN models, with the additional advantage of consuming around 2 times less energy.

Notably, compared with the Iterative LIF neuron model, the proposed NiLIF model improves the accuracy by 0.1 for BCIC IV 2a dataset and 0.2 for OpenBMI dataset respectively, while reducing the energy cost. Table~\ref{tabel_snn_bcic} and \ref{tabel_snn_openbmi} show that the firing rate of NiLIF neurons is much lower than of Iterative LIF neurons, which illustrates the sparsity of our proposed method. The proposed NiSNN and attention mechanisms can be used separately. We present the comparison results of NiSNN and the proposed attention mechanisms on various datasets in the appendix, showing the proposed methods' feasibility.

\subsection{Discussion}
\label{Result.discussion}

\begin{figure}
	\centering
		\includegraphics[width=\linewidth]{ 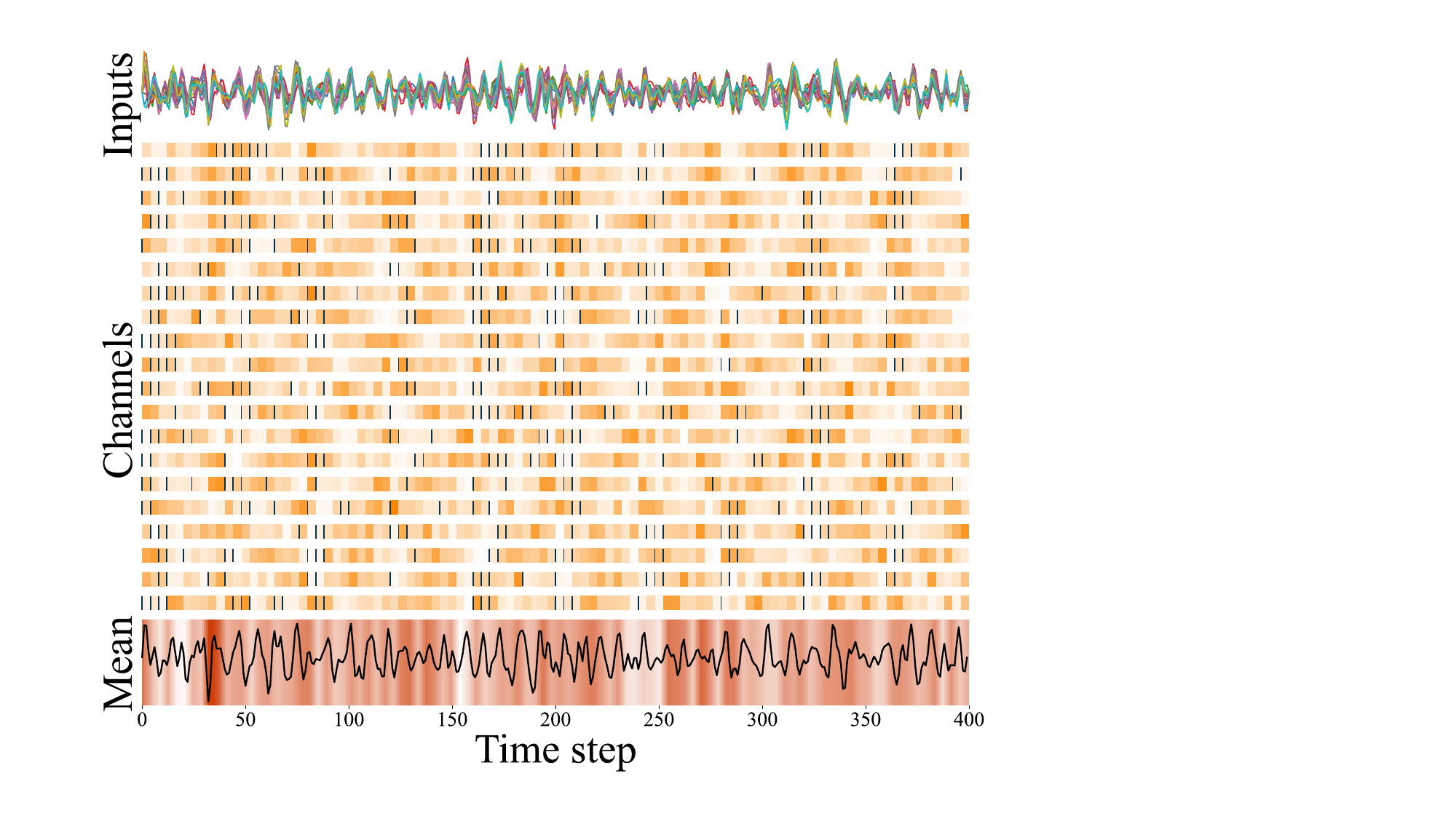}
	  \caption{The generated attention visualization example, using our Global NiSNN-A model. The top figure shows the input EEG signals. In the following figure, black scatters represent the spikes after encoding. The shade of orange blocks represents the degree of attention. The darker orange block shows a higher attention score, and the lighter orange block represents a lower one. In the last figure, the black curve represents the average value of the input EEG on 20 channels.}\label{fig_eeg_spk_attention}
\end{figure}

The experimental results highlight the potential of the attention mechanism in improving the classification accuracy of EEG signals, especially for SNN models. To understand more clearly how the attention mechanism works, we provide a visualization that illustrates the role of the attention score $A$ in the entire feature map. Fig.~\ref{fig_eeg_spk_attention} intuitively represents the attention score $A$ of the Global-attention model.
The top part of the figure depicts the input 20-channel EEG signals in a numeric format. The second part shows the spikes after the spiking encoder, represented by the black raster. The orange blocks represent the attention scores after normalization. Darker orange indicates higher attention scores, while lighter orange indicates lower ones. It illustrates the inner operation of the attention mechanism, highlighting the model's ability to allocate variable attention to different regions of the EEG signal, answering when, where, and what information is relevant.
However, unlike understandable visualizations of attention mechanisms in visual tasks, the physiological meaning of attention scores is not intuitive. Therefore, we show the averaging effect in the third part of the figure. We show the mean of the 20-channel EEG signal, represented by the black line. The orange shading represents higher or lower attention scores. Interestingly, the attention mechanism emphasizes time steps around 30 to 40, corresponding to the most apparent drop in the EEG signal, revealing the importance of this sudden change in the EEG signal.
By employing such a dynamic weighting method, the model ensures that the important regions of the input are prioritized, potentially contributing to the high accuracy in classification tasks.

From an energy consumption point of view, the results highlight the significance of employing SNNs, which achieve accuracy comparable to CNN models and offer a 2-fold reduction in energy use. This efficiency can be important in applications where power consumption is a concern, such as portable EEG devices or real-time EEG monitoring systems. Therefore, SNNs present promising potential for these energy-conscious edge devices.

\section{Conclusion}
\label{Conclusion}

This paper introduced a NiSNN-A model, encompassing NiLIF neurons and diverse attention mechanisms. The newly proposed NiLIF neuron retains the biological attributes of traditional LIF neurons while efficiently handling long temporal data. This design avoids long loops in execution and gradient challenges by approximating the neuron dynamics and calculating synchronously. Subsequently, the attention mechanism emphasizes important parts of the feature map. Notably, all our proposed attention models integrate computations within one singular model instead of using sequential architectures. We employed the BCIC IV 2a and OpenBMI datasets for validation, adopting a subject-independent approach to demonstrate the model's capabilities in uniformed feature extraction for unfamiliar participants. The results indicate that our approach surpasses other SNN models in accuracy performance. It achieves accuracy comparable to its CNN counterparts but improves energy efficiency. Furthermore, our attention visualization results reveal that our model improves the classification task's accuracy and offers deeper insights into EEG signal interpretation. 

This research has provided a way for novel methodologies in EEG classification, focusing on potential cooperation between attention mechanisms and spiking neural network architectures. In the future, as the field of EEG signal processing continues to evolve, our findings require continued innovation and adaptive strategies to address challenges.

\bibliographystyle{IEEEtran}
\bibliography{tnnls_refs.bib}

\clearpage

\appendix

\subsection{Event-based Vision Tasks with NiSNN}

Non-iterative (Ni) LIF model is a general model for the SNNs. In this section, we show the experiment results of using Ni-LIF in the event-based computer vision (CV) tasks.

\begin{figure}[h]
	\centering
		\includegraphics[width=\linewidth]{ 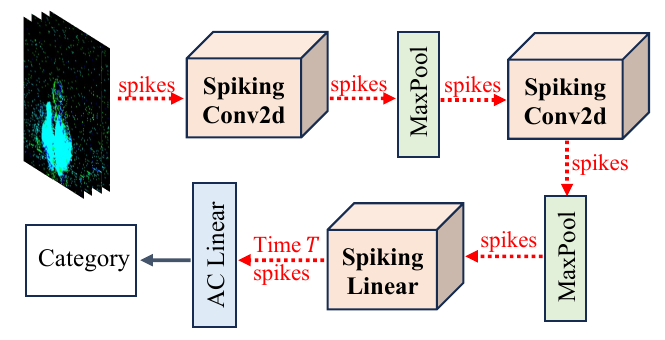}
	  \caption{Overview of the SNN architecture for CV tasks. This SNN comprises two spiking convolutional layers followed by a single spiking linear layer. The network processes input event-based image over multiple time steps, and the spikes generated at the final time step are fed into an output linear layer. The outputs are used for voting to determine the final classification label.}\label{fig_cv_network_structure}
\end{figure}

The overview of the SNN structure is shown in Fig.~\ref{fig_cv_network_structure}. The input event-based image contains $T$ time steps, which is fed into two spiking convolutional layers followed by a spiking linear layer. Finally, a linear layer uses the generated spiking features at the last time step to classify the output label. All network details remain the same for the comparison of Ni-LIF and Iterative LIF models. The experiments use 0.001 as the learning rate and train for 100 epochs.

The comparison takes three event-based datasets: N-MNIST dataset \cite{orchard2015converting}, CIFAR10-DVS dataset \cite{li2017cifar10}, and DVS128 Gesture dataset \cite{amir2017low}. For all three datasets, the input event-based images are resized into 32x32 pixels. The networks are evaluated at three different temporal resolutions, specifically at time steps $T=10$, $T=50$, and $T=100$. The detailed parameters of networks setting is described in Table ~\ref{tab:cv_network_params}.

\begin{table}[h]
\caption{Details of network architecture during training. B is the batch size, and T is the number of time steps.}
\label{tab:cv_network_params}
\begin{tabularx}{\linewidth}{@{} >{\hsize=0.3\hsize}X >{\hsize=0.1\hsize}X >{\hsize=0.2\hsize}X >{\hsize=0.4\hsize}X @{}}
\hline
    Layers & Filters & Size & Output Size\\ 
    \hline
    Input & - & - & (B, 2, 32, 32, T) \\
    Spiking AC Conv2d & 64 & (5, 5) & (B, 64, 28, 28, T) \\
    Max Pooling & - & (2, 2) & (B, 64, 14, 14, T) \\
    Spiking AC Conv2d & 64 & (5, 5) & (B, 64, 10, 10, T) \\
    Max Pooling & - & (2, 2) & (B, 64, 5, 5, T) \\
    Spiking AC Linear & - & - & (B, 11, T) \\
    Linear & - & - & (B, 11) \\
\hline
\end{tabularx}
\end{table}

\begin{figure}[h]
	\centering
		\includegraphics[width=\linewidth]{ 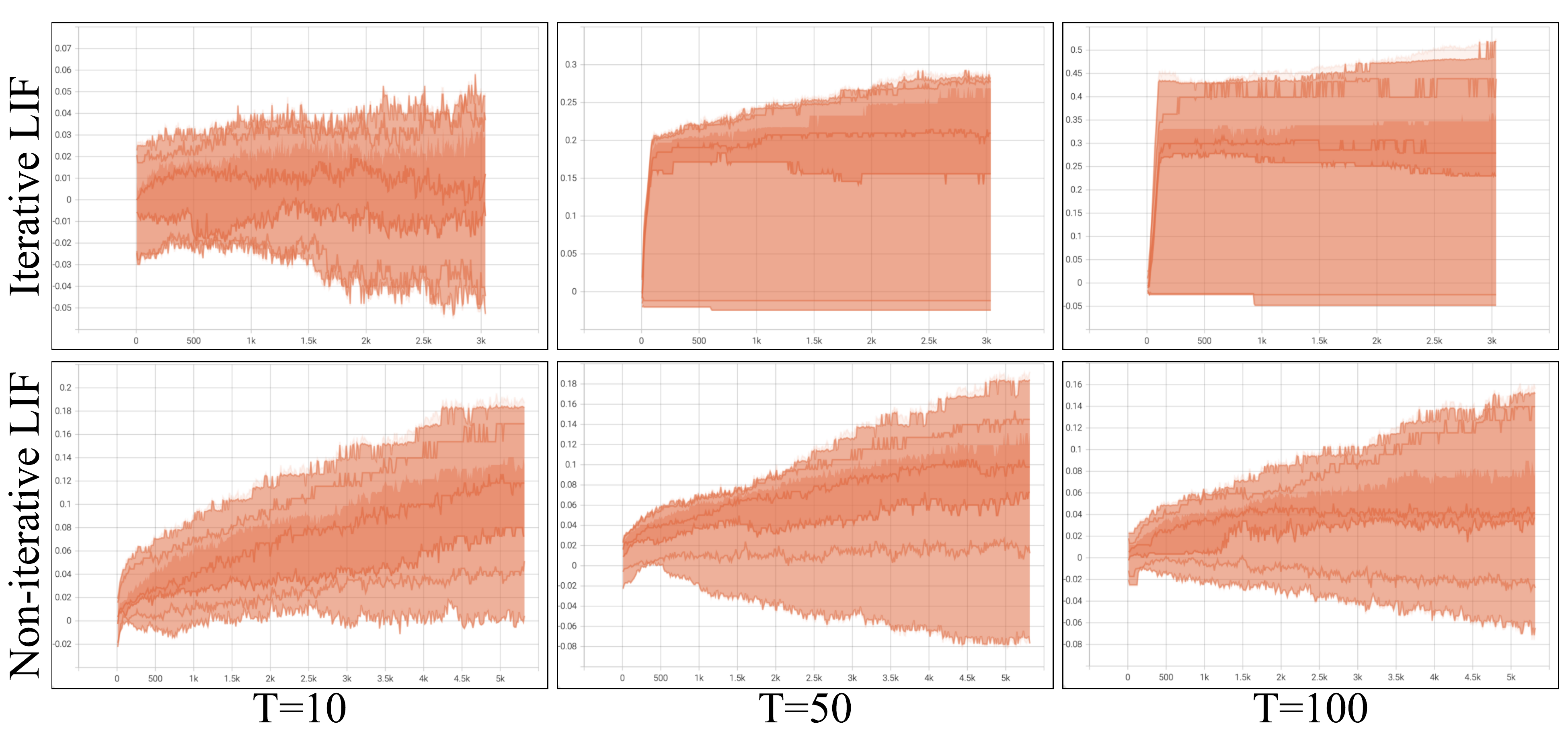}
  \caption{Illustration of gradients in NiSNN and Iterative SNN models for three time-step cases. We show the distribution of the parameters in the Spiking AC Linear layer during the training process using the N-MNIST dataset.}\label{fig_gradient_comparison}
\end{figure}

The experiment results are shown in Table ~\ref{tab:cv_results}. The Ni-LIF model outperforms the Iterative LIF model, and the performance difference is particularly notable at higher time steps (e.g., $T=100$). With the increasing number of time steps, the performance of the Iterative LIF model decreases. 
To illustrate the gradient change straightforwardly, we take the weights distributions in the Spiking AC Linear layer during the training process of the Ni-LIF and Iterative LIF models as an example, which is shown in Fig.~\ref{fig_gradient_comparison}. In the distribution figure, the intensity of the line's color represents the frequency of the corresponding value. When $T=10$, the weights of both models change during training, meaning the gradients bring a usual update of the weights. When $T=50$, there is a plateau in the curve of the Iterative LIF model, which is more obvious when $T=100$. The plateau after the early training epochs represents the minor change of the weights, which implies that small gradients appear in the Iterative LIF model. This phenomenon corresponds to the decreasing performance shown in Table ~\ref{tab:cv_results}. In contrast, the weights of the Ni-LIF model are independent of the time step and vary over time, corresponding to similar performance at all different time steps.

\begin{table}[h]
\caption{Performance Comparison of NiSNN and Iterative SNN in Event-Based Computer Vision Tasks with Varied Time Steps.}
\label{tab:cv_results}
\begin{tabularx}{\linewidth}{@{} >{\hsize=0.4\hsize}X >{\hsize=0.25\hsize}X >{\hsize=0.25\hsize}X >{\hsize=0.15\hsize}X @{}}
\hline
    Dataset & Time Steps $T$ & Neuron Type & Accuracy\\ 
    
    \hline
    \multirow{5}{=}{N-MNIST \cite{orchard2015converting}} & \multirow{1}{*}{10} & Iterative LIF & 0.9781 \\
    ~ &  ~ & Ni-LIF & \textbf{0.9863} \\

    ~ &  \multirow{1}{*}{50} & Iterative LIF & 0.9227  \\
    ~ &  ~ & Ni-LIF & \textbf{0.9868}\\

    ~ &  \multirow{1}{*}{100} & Iterative LIF & 0.7224  \\
    ~ &  ~ & Ni-LIF & \textbf{0.9840}  \\

    \hline
    \multirow{5}{=}{CIFAR10-DVS \cite{li2017cifar10}} & \multirow{1}{*}{10} & Iterative LIF & 0.0866 \\
    ~ & ~ & Ni-LIF & \textbf{0.4588} \\

    ~ & \multirow{1}{*}{50} & Iterative LIF & 0.3505 \\
    ~ & ~ & Ni-LIF & \textbf{0.4458} \\

    ~ & \multirow{1}{*}{100} & Iterative LIF & 0.3180 \\
    ~ & ~ & Ni-LIF & \textbf{0.4508} \\
    
    \hline
    \multirow{5}{=}{DVS128 Gesture \cite{amir2017low}} & \multirow{1}{*}{10} & Iterative LIF & 0.6539 \\
    ~ & ~ & Ni-LIF & \textbf{0.7984} \\

    ~ & \multirow{1}{*}{50} & Iterative LIF & 0.6563 \\
    ~ & ~ & Ni-LIF & \textbf{0.8461}\\

    ~ & \multirow{1}{*}{100} & Iterative LIF & 0.5039 \\
    ~ & ~ & Ni-LIF & \textbf{0.8672} \\
    
\hline
\end{tabularx}
\end{table}

\subsection{Time Series Classification Tasks with NiSNN}

In this section, we show the experiment results of using the Ni-LIF neuron model on other time series classification tasks.

\begin{figure}[h]
	\centering
		\includegraphics[width=\linewidth]{ 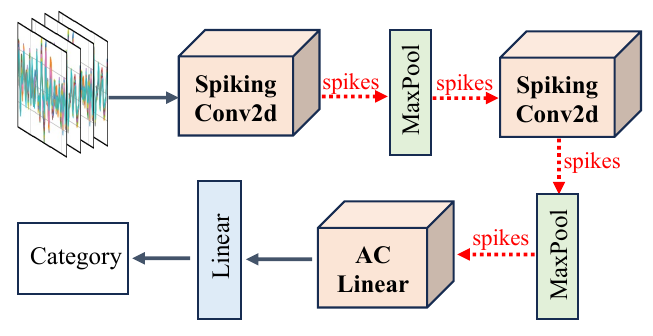}
	  \caption{The SNN architecture for time series classification. This SNN includes two Spiking Convolutional layers, followed by two linear layers. The input time series data are segmented into smaller timepieces, which are processed by the spiking convolutional layers. The output spikes are subsequently fed into the linear layers for determine the final classification label.}\label{fig_ts_network_structure}
\end{figure}

The architecture of the SNN, as depicted in Fig.~\ref{fig_ts_network_structure}, comprises two Spiking convolutional layers followed by two linear layers. These layers facilitate a voting mechanism for determining the final classification label. The network parameters are shown in Table ~\ref{tab:ts_network_params}.

\begin{table}[h]
\captionsetup{labelfont={color=blue}, textfont={color=blue}}
\caption{Details of network architecture during training for time series datasets. B is the batch size, C is the channel size, S is the number of timepieces, and T is the number of time steps in each timepiece.}
\label{tab:ts_network_params}
\begin{tabularx}{\linewidth}{@{} >{\hsize=0.3\hsize}X >{\hsize=0.1\hsize}X >{\hsize=0.3\hsize}X >{\hsize=0.3\hsize}X @{}}
\hline
    Layers & Filters & Size/padding & Output Size\\ 
    \hline
    Input & - & - & (B, C, S, T) \\
    Spiking Conv2d & S & (1, 5)/same & (B, S, S, T) \\
    Max Pooling & - & (2, 2) & (B, S, $\lfloor \frac{S}{2} \rfloor$, $\lfloor \frac{T}{2} \rfloor$) \\
    Spiking AC Conv2d & S & (10, 10)/same & (B, S, $\lfloor \frac{S}{2} \rfloor$, $\lfloor \frac{T}{2} \rfloor$) \\
    Max Pooling & - & (2, 2) & (B, S, $\lfloor \frac{S}{4} \rfloor$, $\lfloor \frac{T}{4} \rfloor$) \\
    AC Linear & - & - & (B, 20) \\
    Linear & - & - & (B, 2) \\
\hline
\end{tabularx}
\end{table}

We take two time-series datasets for the experiments: the Strawberry dataset \cite{holland1998use} from the UCR Time Series Classification Archive \cite{dau2019ucr} and the EEG classification dataset,  BCIC IV 2a \cite{tangermann2012review}. The Strawberry dataset contains the food spectrographs from the strawberry (authentic samples) and non-strawberry (adulterated strawberries and other fruits), which include 613 pieces of training data and 370 pieces of test data. Each piece of data contains 235 time steps. BCIC IV 2a dataset is the EEG dataset for the motor imagery classification task, which contains 9 subjects. Each subject has 288 data records lasting 4 seconds with 20 channels and a sampling frequency of 250 Hz. We downsampled the data to reduce the time steps to 400. For the BCIC IV 2a dataset, we set S as 20 and T as 20. For the Strawberry dataset, we set S as 15 and T as 15.

\begin{table*}[hb!]
\captionsetup{labelfont={color=blue}, textfont={color=blue}}
\caption{Performance comparison of NiSNN and Iterative SNN among time series datasets.}
\label{tab:nilif_ts_results}
\begin{tabularx}{\linewidth}{@{} >{\hsize=0.27\hsize}X >{\hsize=0.2\hsize}X >{\hsize=0.2\hsize}X >{\hsize=0.2\hsize}X >{\hsize=0.2\hsize}X @{}}
\hline
    Dataset & Type  & Time Steps & Neuron Type & Accuracy\\ 
    
    \hline
    \multirow{1}{=}{UCR Strawberry \cite{dau2019ucr}} & \multirow{1}{=}{Spectrograph} &  \multirow{1}{*}{235} & Iterative LIF & 0.8966 \\
    ~ & ~ & ~ & Ni-LIF & 0.9006 \\

    \hline
    \multirow{1}{*}{BCIC IV 2a \cite{tangermann2012review}} & \multirow{1}{*}{EEG} & \multirow{1}{*}{400} & Iterative LIF & 0.5087 +/- 0.0107 \\
    ~ & ~ & ~ & Ni-LIF & 0.6311 +/- 0.1293  \\
    
\hline
\end{tabularx}
\end{table*}

The experiment results are shown in Table~\ref{tab:nilif_ts_results}. The Ni-LIF model outperforms the Iterative LIF model. As with training on the OpenBMI dataset, the training algorithm for the BCIC IV 2a dataset works in a subject-independent manner. The test accuracy is taken as the average of 9 results.

\subsection{Time Series Classification Tasks with Proposed Attention Mechanisms}

The proposed attention mechanisms are general, not only for SNNs but CNNs. In this section, we show the experiment results of using proposed attention mechanisms with CNN for time series datasets. We test on the EEG datasets, BCIC IV 2a and OpenBMI, and the spectrograph dataset, Strawberry from UCR.

\begin{table}[h]
\captionsetup{labelfont={color=blue}, textfont={color=blue}}
\caption{Performance Comparison of proposed attention mechanisms with the vanilla CNN on different time series datasets.}
\label{tab:attention_ts_results}
\begin{tabularx}{\linewidth}{@{} >{\hsize=0.2\hsize}X >{\hsize=0.5\hsize}X >{\hsize=0.3\hsize}X @{}}
\hline
    Dataset & Method  & Accuracy\\ 
    
    \hline
    
    \multirow{5}{=}{BCIC IV 2a} & Vanilla CNN  & 0.6866 +/- 0.0885  \\
    ~ & Our Linear-Seq-attention & 0.6884 +/- 0.1027 \\
    ~ & Our Conv-Seq-attention & \textbf{0.6892 +/- 0.1178} \\
    ~ & Our Linear-ChanSeq-attention & 0.6502 +/- 0.1162 \\
    ~ & Our Conv-ChanSeq-attention & 0.6823 +/- 0.1115 \\
    ~ & Our Global-attention & 0.6658 +/- 0.1087 \\
    
    \hline

    \multirow{5}{=}{OpenBMI} & Vanilla CNN  & 0.7370 +/- 0.1314  \\
    ~ & Our Linear-Seq-attention & 0.7358 +/- 0.1344 \\
    ~ & Our Conv-Seq-attention & \textbf{0.7427 +/- 0.1299} \\
    ~ & Our Linear-ChanSeq-attention &  0.6819 +/- 0.1262 \\
    ~ & Our Conv-ChanSeq-attention & 0.6940 +/- 0.1306 \\
    ~ & Our Global-attention & 0.7412 +/- 0.1304 \\

    \hline

    \multirow{5}{=}{Strawberry UCR} & Vanilla CNN  & 0.9383 +/- 0.0031  \\
    ~ & Our Linear-Seq-attention & 0.9244 +/- 0.0032 \\
    ~ & Our Conv-Seq-attention & 0.9406 +/- 0.0029 \\
    ~ & Our Linear-ChanSeq-attention & 0.9512 +/- 0.0022 \\
    ~ & Our Conv-ChanSeq-attention & 0.9434 +/- 0.0028 \\
    ~ & Our Global-attention & \textbf{0.9516 +/- 0.0027} \\
    
\hline
\end{tabularx}
\end{table}

The network structure follows Fig~.\ref{fig_cnn_structure} and the network parameters are the same as described in Table~\ref{tab_network_structure}. For OpenBMI and BCIC IV 2a datasets, the channel parameter C is set as 20. For the Strawberry dataset, C is equal to 15. The experiment results are shown in Table~\ref{tab:attention_ts_results}. For both EEG datasets, our Conv-Seq-attention mechanism outperforms all other methods. For the Strawberry dataset, our Global-attention mechanism gains the highest accuracy.

These experiments show that the proposed attention mechanisms can be used not only on SNNs but also achieve good results on CNNs among different time series datasets, which demonstrates the generalization.

\vfill

\end{document}